\documentclass{article}

\usepackage[final]{changes}

\usepackage{fancyhdr}
\usepackage[left,pagewise]{lineno}
\usepackage[style=numeric]{biblatex}
\AtEveryBibitem{%
  \clearfield{url}%
  \clearfield{doi}%
  \clearfield{eprint}%
  \clearfield{issn}%
  \clearfield{isbn}
}

\usepackage{graphicx}
\usepackage{subcaption}
\usepackage{multirow}
\usepackage{amssymb}
\usepackage{enumitem}
\usepackage{physics}
\usepackage{makecell}
\usepackage{amsmath}
\usepackage{amsthm}
\usepackage[ruled,vlined, linesnumbered]{algorithm2e}

\usepackage[most]{tcolorbox}
\usepackage{xcolor}

\newcommand{\ig}{\operatorname{IG}}
\newcommand{\ft}{\operatorname{\mathcal F}}
\newcommand{\sgn}{\operatorname{sgn}}
\newcommand{\bvarphi}{\boldsymbol{\varphi}}
\newcommand{\bbbn}{\mathbb{N}}
\newcommand{\bbbr}{\mathbb{R}}
\newcommand{\bbbc}{\mathbb{C}}
\newcommand{\bbbz}{\mathbb{Z}}

\newtheorem{observation}{Observation}
\newtheorem{proposition}{Proposition}
\newtheorem{definition}{Definition}
\newtheorem{lemma}{Lemma}

\begin{document}

\title{Surrogate Modeling for Explainable Predictive Time Series Corrections}
\author{Alfredo L\'opez\thanks{Software Competence Center Hagenberg, Hagenberg, Austria. \texttt{alfredo.lopez@scch.at}} \and Florian Sobieczky\footnotemark[\value{footnote}]}
\maketitle             

\begin{abstract}
We introduce a local surrogate approach for explainable time-series analysis. An initially non-interpretable predictive model to improve the forecast of a classical time-series 'base model' is used. 'Explainability' of the correction is provided by fitting the base model again to the data from which the error prediction is removed (subtracted), yielding a difference in the model parameters which can be interpreted. We provide illustrative examples to demonstrate the potential of the method to discover and explain anomalies by new types of plots showing and comparing the features' importance as a function of time and relative to each other.
\end{abstract}

\section{Introduction}\label{sec:intro}

Explainable AI (XAI) is now a broad and eminent discipline and has seen a tremendous increase of demand, particularly for applications for which accountability of predictive models is crucial \cite{Minh,Dosilovic2018,Carvalho2019,DARPA}. In particular, AI in time series models needs explainable (i.e. interpretable \cite{Burkart}) results in data approximation and prognosis \cite{rojat2021,Ozy22,theissler,guilleme}. {\em Local surrogate modeling} methods such as LIME \cite{LIME,MENG,tsmule,Sivill}, SHAP \cite{SHAP,ksIQ,treeshap} and others \cite{SDM,windowshap} have delivered the possibility to interpret the action of predictive models around a specific instance, that is, in the {\em neighborhood} of a single data point in the input space. The local approach involves solving the problem of choosing the right size and position of the local subset in the incidence space where the surrogate is defined \cite{sundararajan20b,LIMErig,fuzzy,visani,molnar2019}.

To understand this problem in the broader contet of XAI, it is important to note that there are many different aspects to XAI with one common theme: The characterization of the eminence of the learning models' {\em input variables}. Causal inference establishes insight into the probability of an observable (input variable) causing an event \cite{naser}\deleted{\textvisiblespace}, trustworthiness comes from model robustness under changes in the input variable \cite{trustworthy},  and physical interpretations \cite{pinn, raissi19} are characterized by established laws about the influence of measurable parameters. The particular sub-discipline of local surrogate modeling, however, is characterized in terms of using the ideas of sensitivity analysis to explain the importance of single input features $X_i$ of the model's behavior. This means looking at a given local region of the input space surrounding an instance (data point) and describing the shape of the graph locally around it. While PDP \cite{friedman2001} and ALE-plots \cite{ale} do this by observing these changes over single features and aggregating the effects of the other variables, ICE plots \cite{ice} deliver a solution to this to a stronger degree (see e.g. the discussion in \cite{khan25}). The type of data for which the viewpoint of an 'understandable' replacement model - defined only on the neighborhood of single data points - lends insight into the original prediction model necessarily entails the idea of analytical regularity (see \cite{continuity}, Sect. 2.1) and interpolation \cite{interp}. In other words, there must be reasonable hope to understand the shape of the graph of the black-box AI-model in a local neighborhood from knowledge of its shape on close-by instances of the given input data-points. The term 'local surrogate' is appropriate here, as 'surrogate' refers to the interpolating quality between given data points, and 'local' to the restriction to a neighborhood in the input space of available data points. Showing the applicability of the local surrogate concept to time-series data - by adopting the right notion of closeness - is precisely the goal of this paper.

As local model agnostic XAI-methods (see Sect. 7.2.2 in \cite{khan25}) are defined by the attempt to mimic any previously given black box {\em locally} by a different, interpretable model, uncertainties of these surrogate models may play a major role in the quality of the emergent explanations (\cite{consensus}, Sect. 5.1.11). From the viewpoint of defining explainability by local surrogates, it is therefore imperative to discuss {\em fidelity} (of the explaining model toward the one to be explained) in connection with model accuracy. Such a definition has been supplied in \cite{BAPC}. Adopting the surrogate model structure of this work, we focus on explaining the action of a non-interpretable predictive model with higher predictive power (than the base model) in its role as a small {\em additive correction} to an interpretable base model. 

Further approaches yielding explanations of the action of predictive models can be seen in a particular branch of physics-informed, or knowledge-guided machine learning \cite{SelMPJ23}. Also, entropy-based methods are successful in explaining time series anomalies in a physical context \cite{BukovskyKH19}. The approach called "Before and After Prediction Parameter Comparison (BAPC)" \cite{BAPC} has a characteristic way of expressing explanations: \deleted{F}\added{f}irst, the model describing the given data is chosen to be a hybrid model, consisting of an interpretable base model describing the 'bulk' behavior of the data, and an AI-model acting as an additive corrector to the base model. Then, to explain a single instance in the input space, a neighborhood is selected and another fit of the interpretable base model is carried out with data modified (corrected by the AI model to be explained) on just this neighborhood. The change of the interpretable fitting parameters is then taken as the 'explanation' of the local action of the correcting AI model (see Section \ref{sec:bapc}). In this way, the explanation is delivered in terms of the base model, or rather, how it has to be changed (locally), to perform like the initial hybrid model (including the AI-corrector). The change (in the form of different fitted model-parameters) represents what initially the AI-correction model did - hence the name referring to the comparison of parameters 'before' and 'after correction'.

The present work extends the BAPC method to time series modeling, where an interpretable time series model is combined with a correcting highly accurate machine learning model. The latter is explained inside a suitably chosen time window that assumes the role of the neighborhood. Instead of intervals or ball-shaped regions in $\bbbr^d$, we consider  temporal neighborhoods defined by a fixed number of past time steps, analogous to the window-length parameter in \cite{windowshap}.

Our main contribution consists in introducing the BAPC-approach to time series data. This refers to an additive {\em hybrid} model $\widehat{y}(t) = f_\theta(t) + \widehat{\varepsilon}(t)$ consisting of an interpretable base model $f_\theta(t)$, explaining the bulk of the phenomenology of the series by interpretable fitted parameters, and an {\em AI-correction} $\widehat{\varepsilon}(t)$, i.e., a machine learning model, predicting the residual error of the base model, however needing {\em interpretation}. This is delivered in the form of a local surrogate $\Delta f_r=f_{\theta_0}-f_{\theta_r}$ given by the change $\Delta \theta_r =\theta_0 - \theta_r$ of the base model's parameters necessary to {\em locally} produce the observed correction within the fidelity-bounds.  {\em Locally} means on a sub-window of size $r$, where $\Delta \theta_r =\theta_0 - \theta_r$ points into the direction where the correction acts.

We organize this paper as follows: In Section 2 theoretical setup for the concepts of BAPC for time series (SBAPC) is laid out. In Section 3 the connection between a physical and a time series model for discrete time is presented. Section 4 contains the description of another established concept from XAI (the integrated gradient method), and Section 5 contains the experiments - both, on artificial and real data.

\section{Before and After Prediction Parameter Comparison (BAPC)}\label{sec:bapc}
In this section, we specialize the BAPC framework proposed in \cite{BAPC} to the context of explainable time series forecasting. In a three-step process, the base model is first trained on the original time series, resulting in residual errors on which the AI-correction model is subsequently trained. The base model is then retrained on the data corrected by the AI model’s predictions, yielding a parameter change characteristic of this correction. 

For a given real-valued time series $y = (y_t)_{t=1}^n$ of finite length \(n \in \bbbn\), the BAPC consists of the following three steps:

{\noindent \bf Step-1: First application of the base model.} Initially, the BAPC involves fitting a parametric {\em base model} $f_{\theta} : \bbbr \to \bbbr$ on the time series $y$, resulting in
\begin{equation}\label{eq:step1}
	y_t = f_{\theta_0}(t) + \varepsilon_{t}, \, t=1,\ldots,n,
\end{equation}
where  $\theta_0 =(\theta_{01},\ldots,\theta_{0q}) \in R^q$ is the estimated parameter and $ \varepsilon_{t} := y_t - f_{\theta_0}(t)$, $t=1,\ldots,n$, are the residuals.

{\noindent \bf Step-2: Application of the correction model.} The base model chosen in Step-1 is interpretable but lacks overall accuracy, which motivates the use of an additional correction. Therefore, in this step, we apply a {\em correction model}  $g$ to the residuals $\varepsilon_1,\ldots,\varepsilon_n$ obtained in Step-1, leading to the fitted correction model $\widehat{g}$ and predicted residuals $\widehat{\varepsilon}_t := \widehat{g}(t)$,  $t=1,\ldots,n$. The forecast generated by the combined model $f_{\theta_0} + \widehat{g}$ is expected to be accurate, but it will, in general, lack interpretability, prompting the subsequent step.

{\noindent \bf Step-3: Second application of the base model.} In this step, we take a suitable {\em correction window} size $r \in \bbbn_0$ and compute the modified time series $y^{\prime}:=(y^{\prime}_t)_{t=1}^n$ defined as $y^{\prime}_t = y_t$ if $1 \leq t \leq n - r$ and $y^{\prime}_t = y_t - \widehat{\varepsilon}(t)$ if $n-r < t \leq n$. We then fit again the base model $f_{\theta}$ to $y^{\prime}$  
leading to estimated parameter $\theta_r = (\theta_{r1},\ldots,\theta_{rq}) \in \bbbr^q$, the {\em BAPC-explanation}
\begin{equation}\label{eq:deltatheta}
	\Delta \theta_r = (\Delta \theta_{r1}, \ldots,\Delta_{rq}) := \theta_0 - \theta_r \in \bbbr^q
\end{equation}
and the {\em surrogate model} 
\begin{equation}\label{eq:fr}
f_{r} := f_{\theta_0} + \Delta f_r,
\end{equation}
where $\Delta f_r := f_{\theta_0} - f_{\theta_r}$ is the {\em surrogate correction}. \\

The BAPC procedure is designed to separate forecasting performance from interpretability. The correction model $\widehat{g}$ can be chosen flexibly to improve accuracy, even if it lacks transparency. To explain its effect, we reapply the interpretable base model to a corrected subset of the time series. The difference in parameters $\Delta \theta_r$ then acts as a local surrogate explanation of the predictive correction in terms of interpretable changes.

Note that other concepts than BAPC use the difference of two model fits (notion of the {\em sensitivity axiom} of \cite{caret}, Sect. 5.2), but - to the best of our knowledge - not in combination with BAPC's hybrid model structure of base model and correction model.

\begin{figure}[ht]
\captionsetup[subfigure]{labelformat=empty} 
  \centering
  \begin{subfigure}[t]{0.3\textwidth}
    \includegraphics[width=\linewidth]{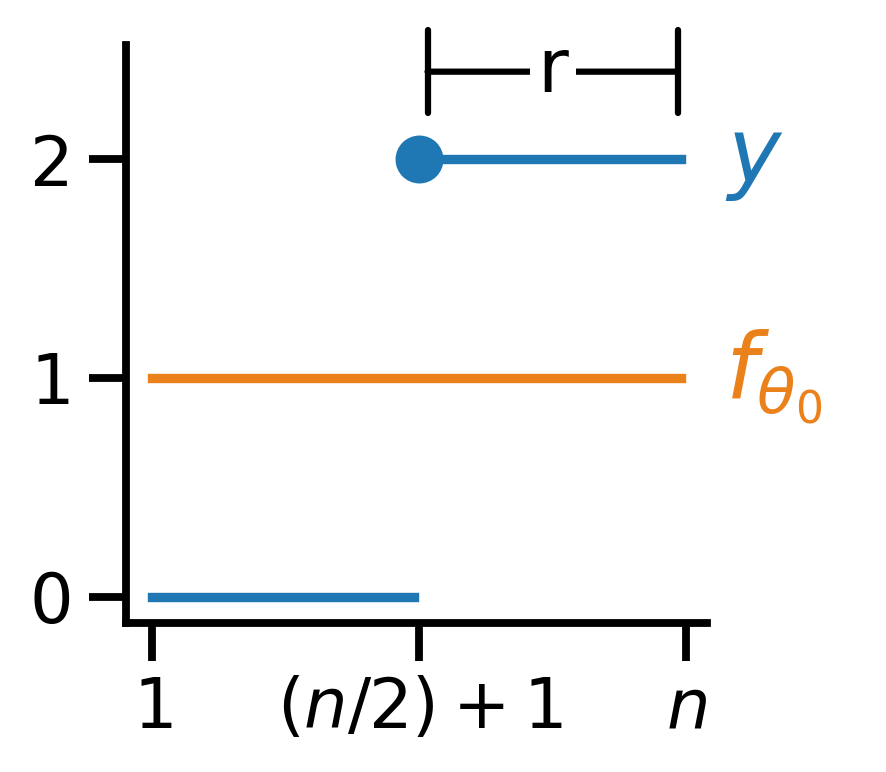}
    \caption{{\bf Step-1}:First application \\ of the base model.}
    \label{fig:BAPC}
  \end{subfigure}
  \begin{subfigure}[t]{0.3\textwidth}
    \includegraphics[width=\linewidth]{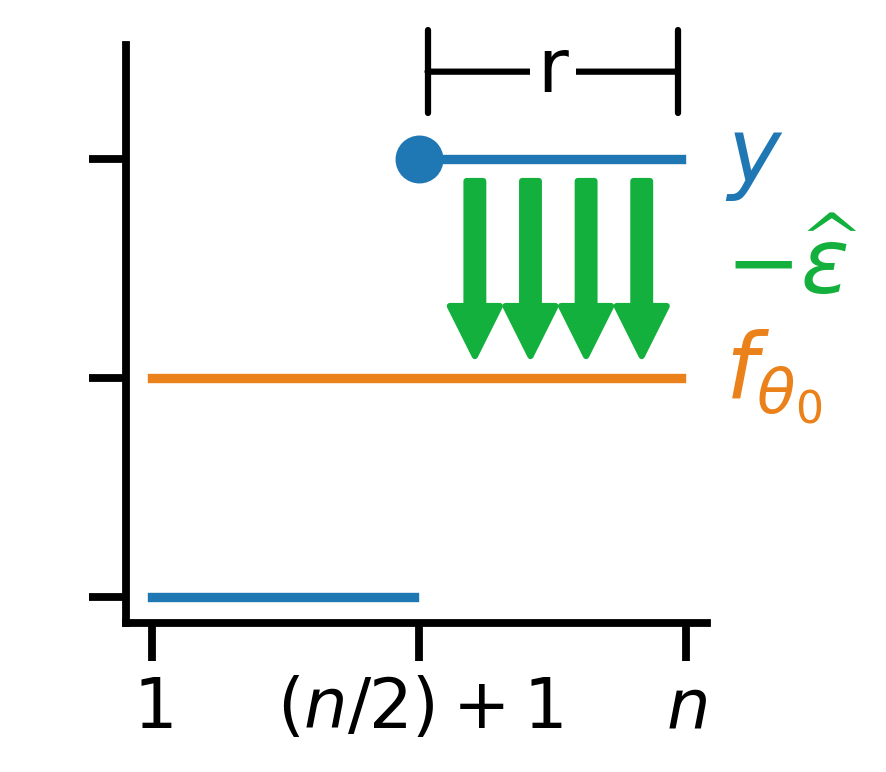}
    \caption{{\bf Step-2}: Application of  \\the correction model.}
    \label{fig:bapc2}
  \end{subfigure}
  \begin{subfigure}[t]{0.3\textwidth}
    \includegraphics[width=\linewidth]{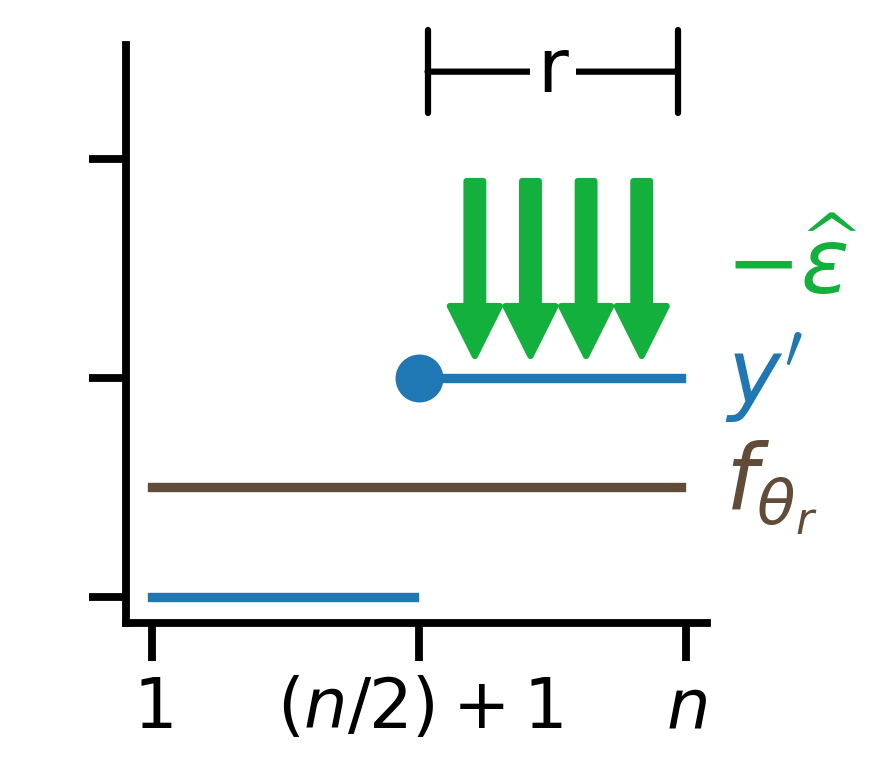}
    \caption{{\bf Step-3}: Second application of the base model.}
    \label{fig:bapc3}
  \end{subfigure}
  \caption{The BAPC applied to the piecewise constant time series $y=2u$.}
  \label{fig:bapc}
\end{figure}

The BAPC is outlined as pseudo-code in Algorithm \ref{alg:bapc} and illustrated through its application to the piecewise constant time series $y$, included as comments in the pseudo-code. The input time-series is given by $y=2u$, where $u$ is a piecewise constant time-series of even length $n$ featuring a unit "jump" at position $(n/2)+1$, namely
\begin{equation*}
u_t = 
\begin{cases}
0, & \added{t} < \frac{n}{2} + 1 \\
1, & \added{t}  \geq \frac{n}{2} + 1,
\end{cases}
\end{equation*}
with $t=1,\ldots,n$. Initially, in Step-1, a constant function $f_\theta(t) = \theta$ is used as the base model, leading to $\theta_0 = 1$ (see line 1 in Algorithm \ref{alg:bapc} and Step-1 in Figure \ref{fig:bapc}). In Step-2 (line 3), we apply the 1-nearest-neighbor interpolation $g$, leading to predicted residuals $\hat{\varepsilon}$ (represented by the vertical arrows in Step-2 of Figure \ref{fig:bapc}). During Step-3, using a window size $r = n/2$, the modified time series $y'$ has a jump of size 1 at $(n/2)+1$ (lines 4-7 and Step-3 of  Figure \ref{fig:bapc}). Consequently, the corrected parameter becomes $\theta_r = 0.5$ (line 8), providing an explanation $\Delta\theta_r = \theta_0 - \theta_r = 1 - 0.5 = 0.5$ (line 9), which points in the (upward) direction of $\widehat{\varepsilon}$ within the correction window. The surrogate model in the constant function $f_r=1+0.5 = 1.5$ which surrogates the complete model $f_{\theta} + \widehat{g}$.

\begin{algorithm}[H]
\DontPrintSemicolon
\SetKwFunction{FitBaseModel}{FitBaseModel}
\SetKwFunction{FitCorrectionModel}{FitCorrectionModel}
\SetKwInOut{Input}{input}
\SetKwInOut{Output}{output}
\SetKwInOut{PhantomInput}{\phantom{input}}
\SetKwInOut{PhantomOutput}{\phantom{output}}
\SetKwFunction{BAPC}{BAPC}
\SetKwProg{Fn}{Function}{:}{\KwRet}
\SetKwProg{Proc}{Procedure}{:}{\KwRet}
    \tcp*{Example usage}
    \Input{$y$ - time series \tcp*{$y=2u$}}
    \PhantomInput{$f_\theta$ - base model \tcp*{$f_\theta = \theta$, $\theta \in \mathbb{R}$}}
    \PhantomInput{$g$  - correction model \tcp*{$g$ = 1-nearest-neighborhood}}  
    \PhantomInput{$r$ - correction window \tcp*{$r=n/2$}}
    \Output{$\Delta\theta_r$ - explanation} 
    \PhantomOutput{$f_r$ surrogate model}
    $\theta_0 \gets$ \FitBaseModel{$f_\theta$, $y$} \tcp*{$\theta_0 = 1$}
    $\varepsilon \gets y - f_{\theta_0}(t)$ \tcp*{$\varepsilon = 2u - 1$}
    $\hat{\varepsilon} \gets$ \FitCorrectionModel{$g$, $\varepsilon$} \tcp*{$\hat{\varepsilon}=\varepsilon=2u-1$ (perfect fit)}
    $n \gets \text{length}(y)$\;
    $y' \gets y$\;
    \For(\tcp*[f]{$y'= u$}){$t \in \{n - r + 1, \ldots, n\}$}{
        $y'[t] \gets y[t] - \hat{\varepsilon}(t)$ \; 
    } 
    $\theta_r \gets$ \FitBaseModel{$f_\theta$, $y'$} \tcp*{$\theta_r = 0.5$}
    $\Delta\theta_r \gets \theta_0 - \theta_r$ \tcp*{$\Delta\theta_r = 1 - 0.5 = 0.5$}
    $\Delta f_r \gets f_{\theta_0} - f_{\theta_r}$ \tcp*{$\Delta f_r= 0.5$ }
    $f_r \gets f_{\theta_0} + \Delta f_r$ \tcp*{$f_r = 1 + 0.5 = 1.5$}

    \BlankLine
    \Fn{\FitBaseModel{$f_\theta$, $y$}}{
        Fit $f_\theta$ to the time series $y$\;
    \KwRet $\theta$ \tcp*{return estimated parameter}
    }
    \Fn{\FitCorrectionModel{$g$, $\varepsilon$}}{
        Fit $g$ to the time series $\varepsilon$\;
    \KwRet $\hat{\varepsilon}$ \tcp*{return predicted values}
    }
    
\caption{BAPC}
\label{alg:bapc}
\end{algorithm}

% Strictly speaking, one would have to call $\Delta f_r$ the surrogate because it surrogates the correction $\widehat{\varepsilon}$.  However, $f_{r}$ surrogates the complete model $f_{\theta} + \widehat{\varepsilon}$, which is of primary interest. Figure \ref{fig:bapc} illustrates the BAPC in a piecewise-constant time series.

\begin{figure}
	\centerline{\includegraphics[width=0.9\textwidth]{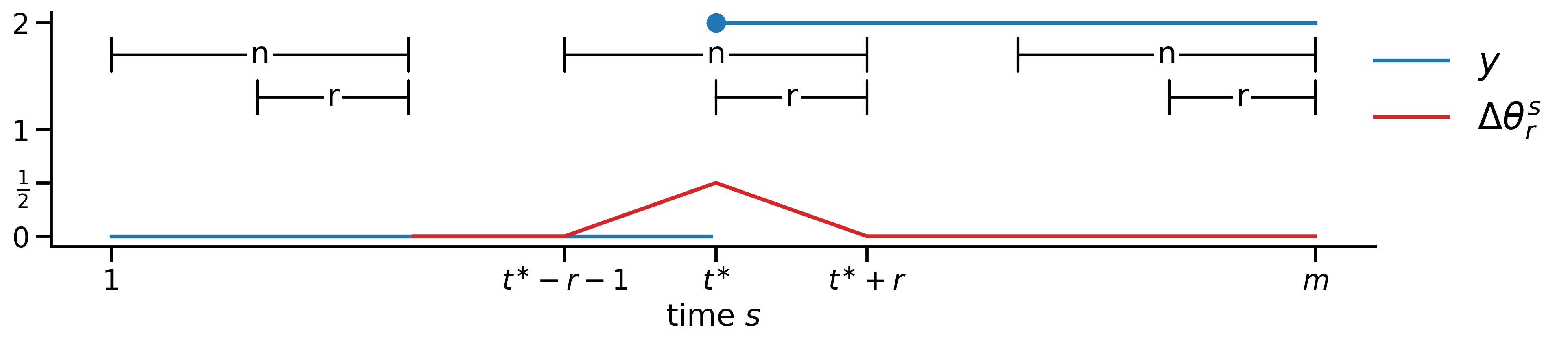}}
	\caption{The sequential-BAPC is applied to a piecewise constant step function (blue) having a discontinuity at time $t^{\ast}$, in a similar setting as in Figure \ref{fig:bapc}, using the constant function as the base model, 1-nearest-neighborhood interpolation as the correction model, and a correction window size $r = n/2$ with even $n$. The operation on the sliding window  with end-points $s-n$ and $s$, for $s=n,\ldots,m$ (horizontal bracketed lines), leads to the SBAPC-explanation $\Delta \theta_r^s$ (red).}
	\label{fig:sbapc}
\end{figure}

For the observations $y_1,y_2, \ldots, y_m$ of a time series taking place consecutively over time, it is of interest to apply the BAPC sequentially for each point in time using the most recent data of a slinding window. We define sequential BAPC (SBAPC) as the process of consecutively applying BAPC to $y_{s-n+1},\ldots,y_s$, for every $s=n,\ldots,m$, where $1 \leq n \leq m$ is a fixed training set size. The BAPC-explanation at time $s$ is given by
\begin{equation}\label{eq:thetatr}
	\Delta \theta_r^s := (\Delta \theta_{r1}^s, \ldots,\Delta \theta_{rq}^s) = \theta_0^s - \theta_r^s \in \bbbr^q
\end{equation}
where $\theta_0^s = (\theta_{01}^s,\ldots,\theta_{0q}^s)$ and $\theta^s_r = (\theta_{r1}^s,\ldots,\theta_{rq}^s)$ are the parameters obtained, respectively, after Step-1 and Step-3 of BAPC applied to $y_{s-n+1},\ldots,y_s$. The surrogate model at time $s$ is $f_{r}^s := f_{\theta_0^s} + \Delta f_r^s$, where $\Delta f_r^s := f_{\theta_0^s} - f_{\theta_r^s}$ is the surrogate correction at time $s$. The SBAPC employs a sliding window approach, wherein an $n$-size training window is moved step by step across the dataset to derive explainability from local sections of the time series. The correction window is defined as the most recent  $r$ points within the training window. The SBAPC is presented as pseudo-code in Algorithm \ref{alg:sbapc} and depicted in Figure \ref{fig:sbapc}.

\begin{algorithm}[H]
\DontPrintSemicolon
\SetKwFunction{FitBaseModel}{FitBaseModel}
\SetKwFunction{FitCorrectionModel}{FitCorrectionModel}
\SetKwInOut{Input}{input}
\SetKwInOut{Output}{output}
\SetKwInOut{PhantomInput}{\phantom{input}}
\SetKwInOut{PhantomOutput}{\phantom{output}}
\SetKwFunction{BAPC}{BAPC}
\SetKwProg{Fn}{Function}{:}{\KwRet}
\SetKwProg{Proc}{Procedure}{:}{\KwRet}

    \Input{$y$ - time series \tcp*{time series of length $m$}}
    \PhantomInput{$f_\theta$ - base model}
    \PhantomInput{$g$  - correction model}  
    \PhantomInput{$r$ - correction window}
    \PhantomInput{$n$ - training set size}
    \Output{$(\Delta\theta_r^s)_{s=n}^m$ - explanations at time $s=n,\ldots,m$} 
    \PhantomOutput{$(f_r)_{s=n}^m$ surrogate models  at time $s=n,\ldots,m$ }

    $m \gets \text{length}(y)$\;
    $\Delta\theta_r \gets [\;]^{m-n}$ \tcp*[r]{Empty list of size $m-n$}
    $f_r \gets [\;]^{m-n}$ \tcp*[r]{Empty list of size $m-n$}
    \For{$s \in \{n, \ldots, m\}$}{
         $\Delta\theta_r^s, f_r^s \gets$ \BAPC{$y[s-n+1:n]$, $f_\theta$, $g$, $r$} \; 
         $\Delta\theta_r[s] \gets \Delta\theta_r^s$ \;
         $f_r[s] \gets f_r^s$ \;   
    } 
\caption{Sequential-BAPC}
\label{alg:sbapc}
\end{algorithm}

\section{Base model}\label{sec:basemodel}

As a base model, we use a function describing decaying oscillations
%\sum_{k=1}^q c_k t^{k-1} + 
\begin{equation}\label{eq:basemodel}
    f_{\theta}(t) = \alpha \exp(-\beta t) \cos(w t + \phi),
\end{equation}
with $\theta=(\alpha,\beta,w,\phi)$, $\alpha, w \in \bbbr_+$, $\beta \in \bbbr$ and $\phi \in [0, 2\pi]$, representing amplitude, logarithmic decrement, frequency, and phaseshift, respectively. The function \eqref{eq:basemodel} is the solution of the simple harmonic oscillator \cite{chaudhuri2001waves}, a model that plays an important role in time series analysis, particularly in understanding periodic behavior, modeling oscillations, and analyzing dynamical systems.  Another related formulation that we consider in this paper is the deterministic auto-regressive sequence $\mathrm{AR(2)}$, which is a particular case of the widely used ARIMA models \cite{lutkepohl2005new}. We say that $y = (y_t)_{t=1}^n$ is a $\mathrm{AR(2)}$ sequence with  coefficients $\varphi_1, \varphi_2 \in \bbbr$ if
\begin{equation}\label{eq:ar}
y_t = \varphi_1 y_{t-1} + \varphi_2 y_{t-2}, \; t=3,\ldots,n. 
\end{equation}

Observe that models \eqref{eq:basemodel} and \eqref{eq:ar} are fully characterized by $(\alpha,\beta,w,\phi)$ and $(y_1,y_2,\varphi_1, \varphi_2)$ respectively. The following two propositions show the connection between these two models; the proofs are postponed to the Appendix \ref{sec:proof}.

\begin{proposition}\label{prop:sin2ar}
 Let $y_t = \alpha \exp(-\beta (t-1))\cos(\omega (t-1) + \phi)$ for $t=1,\ldots,n$, with $\alpha, w \in \bbbr_+$, $\beta \in \bbbr$ and $\phi \in [0, 2\pi]$. Then $(y_t)_{t=1}^n$ is a $\mathrm{AR(2)}$ sequence with $y_1 = \alpha \cos(\phi)$, $y_2 = \alpha \exp(-\beta)\cos(\omega + \phi)$, $\varphi_1 = 2 \exp(-\beta) \cos(\omega)$ and $\varphi_2 = -\exp(-2 \beta)$.
\end{proposition}

\begin{proposition}\label{prop:ar2sin}
Let  $(y_t)_{t=1}^n$ be a $\mathrm{AR(2)}$ sequence with  coefficients $\varphi_1 \in \bbbr$, $\varphi_2<0$, then 
$y_t = \alpha \exp(-\beta (t-1))\cos(\omega (t-1) + \phi)$ for $t=1,\ldots,n$, with $\beta = - \frac{1}{2} \ln(-\varphi_2)$, $\omega = \arccos\left(\frac{1}{2}  \varphi_1  \exp(\beta) \right)$, $\phi = \atan(\frac{y_1 \exp(-\beta) \cos \omega -y_2}{y_1\exp(-\beta) \sin \omega})$ and $\alpha = y_1\sqrt{1 + \tan^2 \phi }$.
\end{proposition} 

Being able to switch between (\ref{eq:basemodel}) and (\ref{eq:ar}) allows obtaining physical interpretations of discrete time series data specific to explanations in the sense of SBAPC. In Proposition \ref{prop:combform} we provide a characterization of the $\mathrm{AR(2)}$ process dynamic in dependence of the initial conditions that is explicit in time and in the auto-regressive coefficients.

We emphasize that the base model must be specified in advance, as it defines the interpretable structure against which deviations are assessed. Its selection relies on prior knowledge about the expected behavior of the time series and the nature of possible anomalies. This introduces a modeling trade-off between interpretability and flexibility, which becomes especially relevant in online settings. We revisit this point in the experimental section, where we test BAPC under mixed anomaly types, including oscillatory and step changes.

\section{Explanation importance}
The SBAPC explanation, $\Delta \theta_r$, defined in Section 2, is a vector of parameter changes of which each coordinate is designed to capture the respective feature's influence on the correction model's predictions. In this section, we introduce the SBAPC Integrated Gradient Score, an adaptation of the Integrated Gradients method \cite{Sundararajan2017,Lundstrom2022ARS}, which aims to quantify and compare the importance of these explanations. Throughout this discussion, we retain the notation and definitions established in Sections \ref{sec:bapc} and \ref{sec:basemodel}. We assume that the mapping $\theta \mapsto f_\theta(t)$ is differentiable for every $t \in \bbbr$, and we denote the partial derivative of $\theta \mapsto f_\theta(t)$ with respect to $\theta_k$ at $\theta' \in \bbbr^q$ by $\frac{\partial f}{\partial \theta_k}(t, \theta')$.

\begin{definition}
    The segment joining $a \in \bbbr^n$ and $b \in \bbbr^n$ is defined by the function $\gamma:\bbbr^{2n}\times[0,1]\to\bbbr^n, h  \mapsto \gamma(a,b,h) := a + h(b-a)$.
\end{definition}

\begin{definition}\label{def:Itheta}
The BAPC integrated gradient of the $k$-th explanation at $t \in \bbbr$ is defined as
    \begin{equation}
        \ig_k(f_{\theta}, t)  :=  \Delta \theta_{rk} \int_0^1  \frac{\partial f}{\partial \theta_k}(t, \gamma(\theta_r, \theta_0, h)) \dd{h}.
    \end{equation}
For the sequential setting with $m$ observations, we define the  SBAPC integrated gradient of the $k$-th explanation at $ s \in \{ n,\ldots,m \}$ and $t \in \bbbr$ as
\begin{equation}\label{eq:igst}
    \ig_k(f_{\theta}, s, t) :=  \Delta \theta_{rk}^s \int_0^1  \frac{\partial f}{\partial \theta_k}(t,  \gamma(\theta_r^s, \theta_0^s, h)) \dd{h}.
\end{equation}
We will also denote $\ig(f_{\theta}, s) := (\ig_1(f_{\theta}, t),\ldots,\ig_1(f_{\theta}, q))$ and $\ig(f_{\theta}, s, t) = (\ig_1(f_{\theta},s, t),\ldots,\ig_q(f_{\theta},s,t))$.
\end{definition}

\begin{observation}\label{obs:ig}
    In the sequential setting, $\ig(f_{\theta}, s, t)$ corresponds to the importance of past predictions, current point in time prediction or future forecast  depending on weather $t<s$, $t=s$ or $t>s$ respectively.
\end{observation}

The following result is a direct consequence of the completeness of Integrated Gradients.

\begin{proposition}\label{prop:completeness}
 The sum of the BAPC integrated gradients components is equal to the correction surrogate, namely
  \begin{equation}
    \sum_{k=1}^{q} \ig_k(f_{\theta}, t)   = \Delta f_r(t).
 \end{equation} 
\end{proposition}
\begin{proof}
    The result follows directly from Proposition 1 in \cite{Sundararajan2017} and the Fundamental Theorem of Calculus.
\end{proof}

We note that the use of a {\em gradient} in the definition of 'explainable' is characteristic of local methods \cite{continuity, khan25} for which there is data with sufficient regularity in time. To be able using this concept for time-series data relies on the ability to switch from the continuous time representation (5) and its discrete time variant (6). We note a related approach to local surrogate modeling using evaluation of time shifts \cite{nakano} 

In Proposition~\ref{prop:ig_formulas} in the appendix, we derive expressions for $\ig(f_{\theta}, t)$ in the case of a linear base model, as well as for the base models considered in Section~\ref{sec:basemodel}. These formulas provide explicit expressions that offer both analytical insight and a foundation for numerical implementation.

\section{Experiment and analysis}\label{sec:experiments}

In this section, we demonstrate and evaluate the proposed method using four synthetic examples, presented in increasing order of complexity, followed by two real-world case studies \footnote{Numerical implementation at \url{https://github.scch.at/lopez/pitsa}.}

\subsection{Trend change point}\label{sec:tren}

\begin{figure}
  \centering
  % First row
  \begin{subfigure}[b]{0.25\textwidth}
    \centering
    \includegraphics[width=\textwidth]{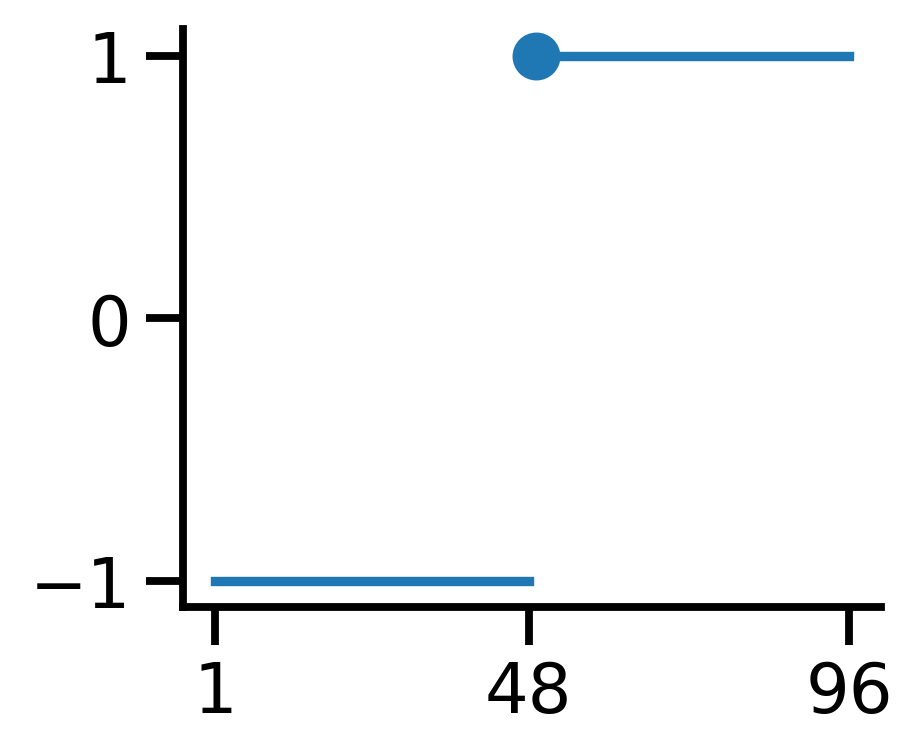}
  \end{subfigure}%
  \begin{subfigure}[b]{0.25\textwidth}
    \centering
    \includegraphics[width=\textwidth]{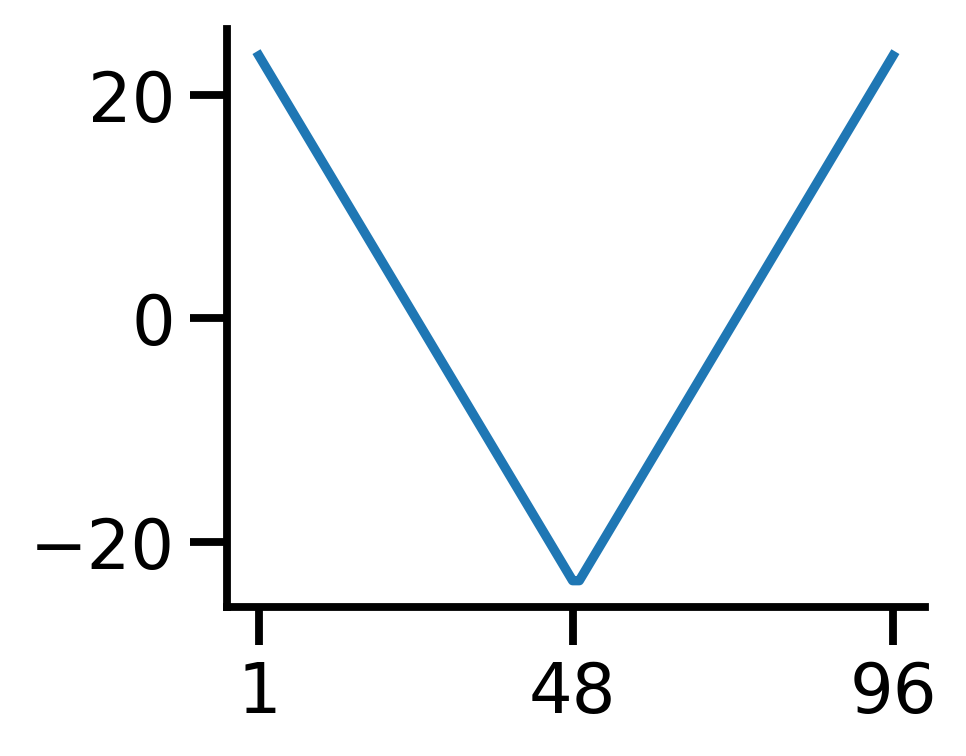}
  \end{subfigure}%
  \begin{subfigure}[b]{0.25\textwidth}
    \centering
    \includegraphics[width=\textwidth]{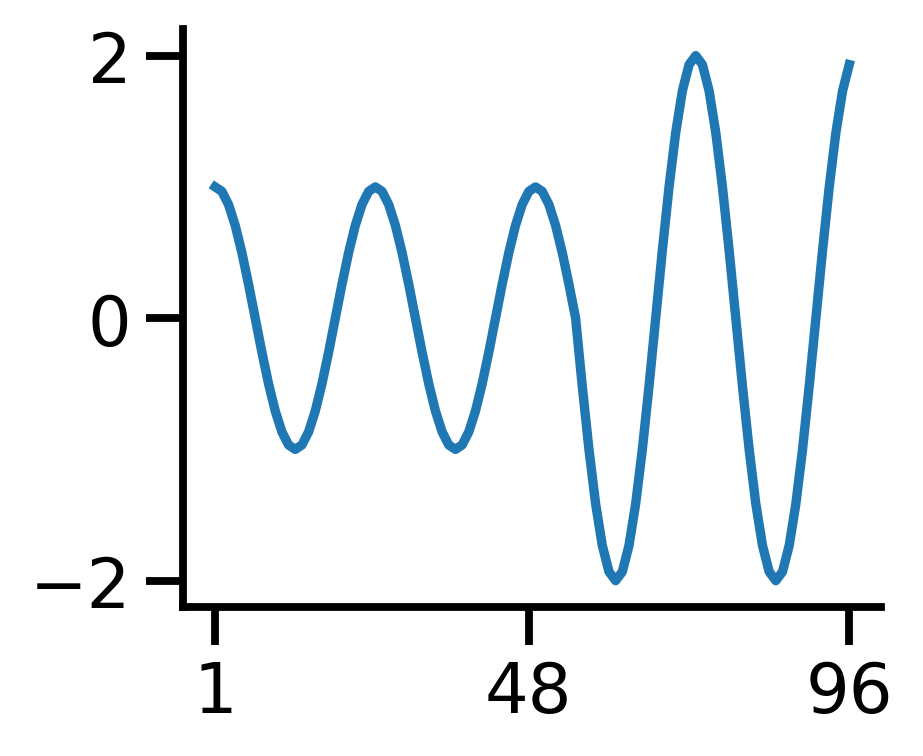}
  \end{subfigure}
   \begin{subfigure}[b]{0.15\textwidth}
    \centering
    \raisebox{30pt}{\includegraphics[width=\textwidth]{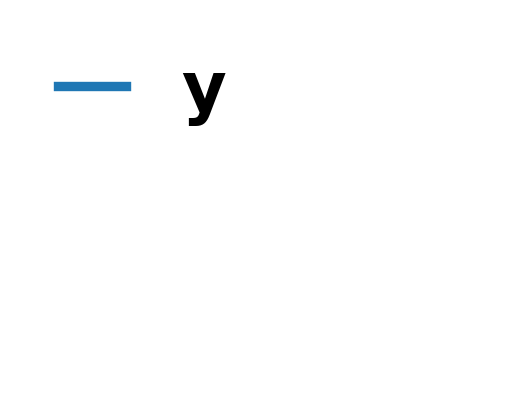}}
  \end{subfigure}
  
  % Second row
  \begin{subfigure}[b]{0.25\textwidth}
    \centering
    \includegraphics[width=\textwidth]{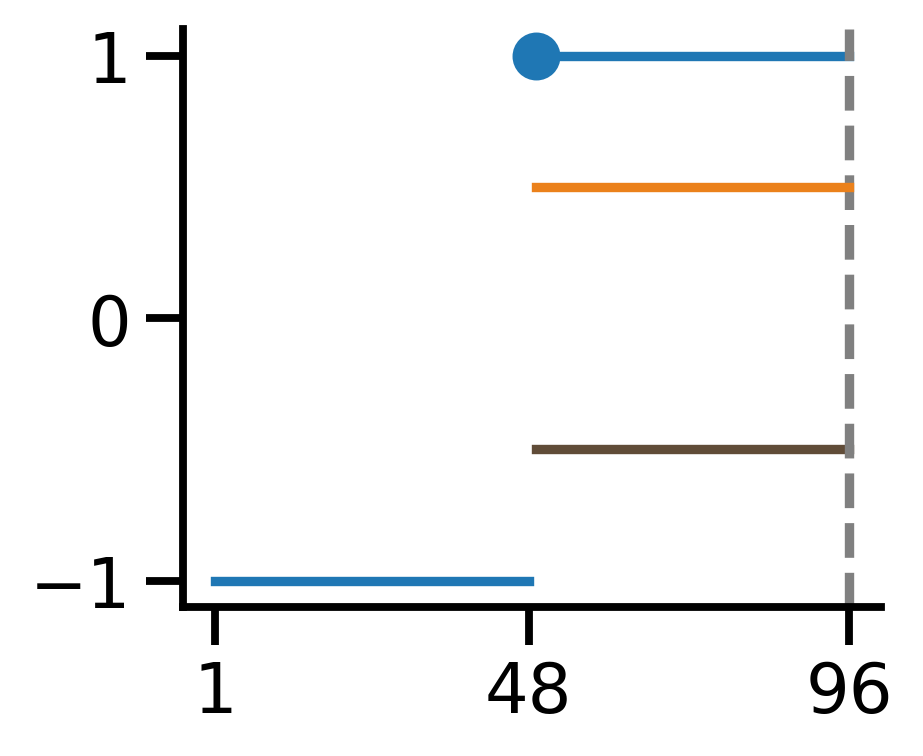}
    \caption{}
    \label{fig:syn1}
  \end{subfigure}%
  \begin{subfigure}[b]{0.25\textwidth}
    \centering
    \includegraphics[width=\textwidth]{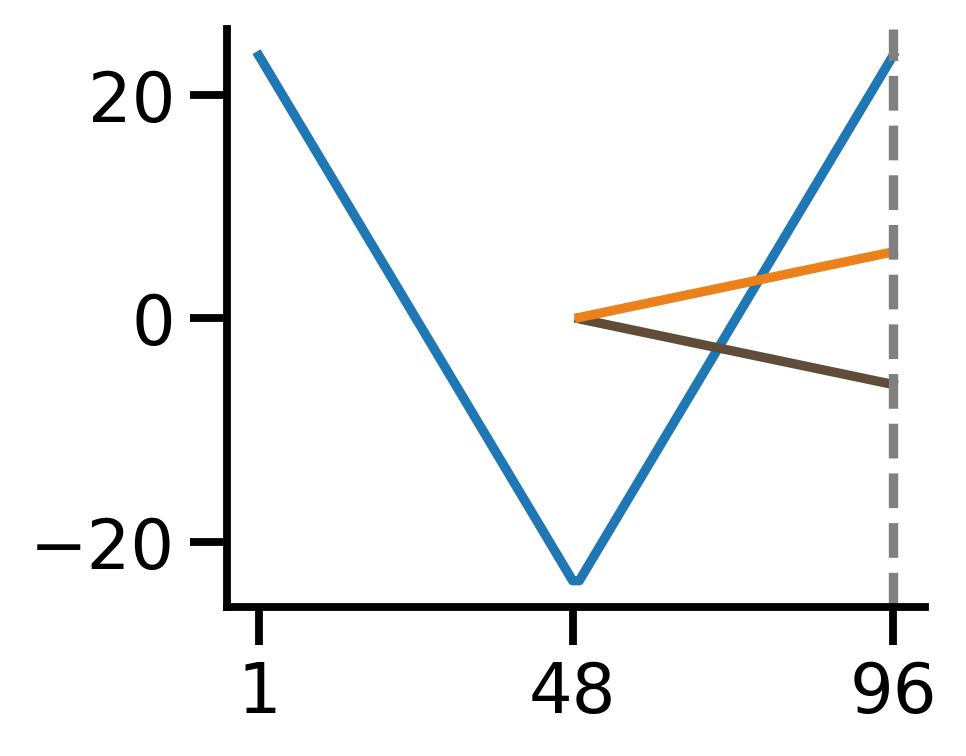}
    \caption{}
    \label{fig:syn2}
  \end{subfigure}%
  \begin{subfigure}[b]{0.25\textwidth}
    \centering
    \includegraphics[width=\textwidth]{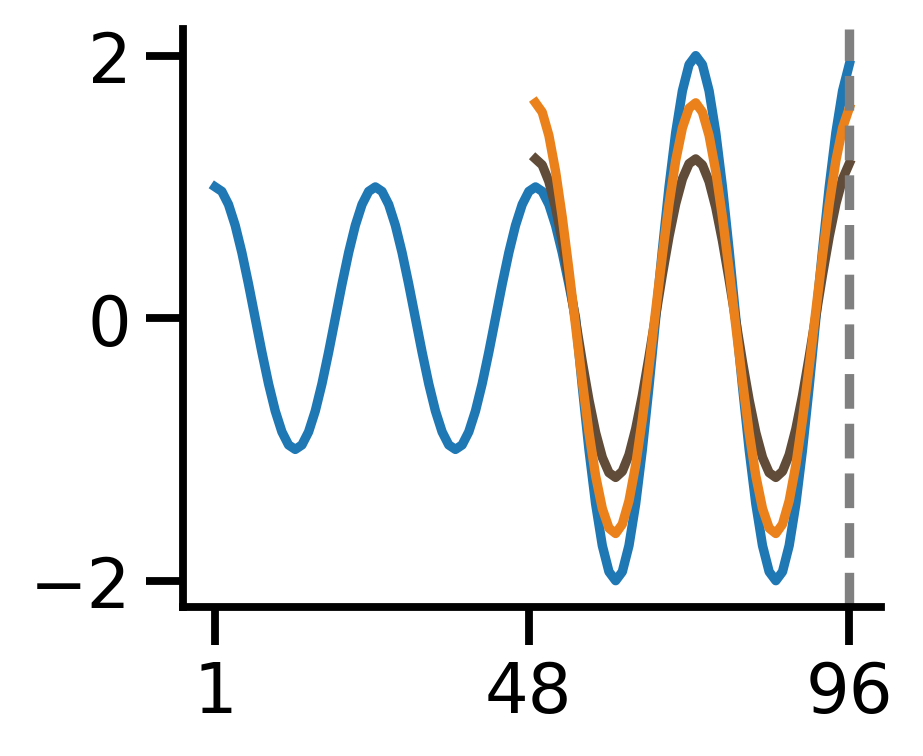}
    \caption{}
    \label{fig:syn3}
  \end{subfigure}
  \begin{subfigure}[b]{0.15\textwidth}
    \centering
    \raisebox{36pt}{\includegraphics[width=\textwidth]{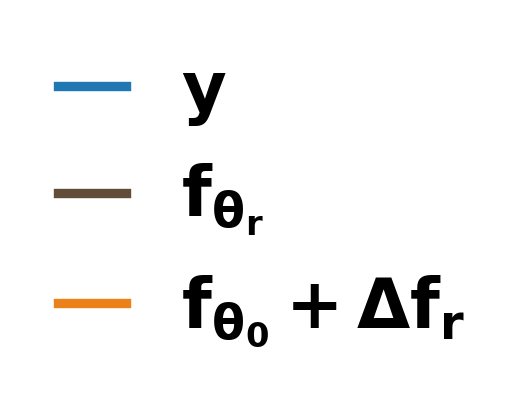}}
    \label{fig:d}
  \end{subfigure}
  \caption{Experimental evaluation on (a) step, (b) ramp and (c) sinusoidal data. Each time series has length $n=96$ and a change point at $t=49$ associated to (a) intercept, (b) intercept and slope and (c) amplitude and phase. The surrogate of the hybrid model is shown by the orange curves: For (a), an increase in the intercept, for (b) an increase in the slope together with a decrease in the intercept, and for (c) a growing in amplitude with an small phase adjustment.}
  \label{fig:synthetic1}
\end{figure}

The first row in Figure \ref{fig:synthetic1} displays three synthetic time series corresponding to (a) step, (b) ramp, and (c) sinusoidal patterns. In  Appendix \ref{sec:ode} we show that these data correspond to discrete observations from one dimensional dynamical systems governed by a non-homogeneous ODE. Each of these time series has a length of $n=96$ and a change point at $t=49$, associated respectively with (a) intercept, (b) intercept and slope, and (c) amplitude and phase. For each of these input time series, we carry out BAPC with a correction window size of $r=40$ and the base model $f_{\theta}$ described in the third column of Table \ref{tab:experiment}. For correction model we take the 1-Nearest-Neighbors regression, which is well suited for the current illustrative example as generate perfect fit on train data. On the other hand, the 1-Nearest-Neighbors is very prone to overfitting, hence in oncoming examples we will consider an actual AI model option. The second row in Figure \ref{fig:synthetic1} shows the base model after correction (brown line $f_{\theta_r}$) and the surrogate model (orange line $f_0 + \Delta f_r$). As expected, the surrogate model aligns with the direction of the change in the data within the correction window relative to the remaining training window. 

From Proposition \ref{prop:ig_formulas}-(1)-(2), we obtain the expression for $\ig(f_\theta, t)$ shown at the fourth column of Table \ref{tab:experiment}, where $\Delta g_r (t) = \sin(\omega t + \phi_0) - \sin(\omega t + \phi_r)$. The last two columns show the explanation $\Delta \theta_r$ and integrated gradient $\ig(f_\theta, t)$ at $t=96$. From these values, we conclude that the method effectively explain the correction model prediction and generates a meaningful feature importance ranking of these explanations in terms of the base model parameters.

\begin{table}[ht]
    \centering
    \caption{Experimental configuration and results for step, ramp and sinusoidal with amplitude change point (SinACP) data.}
    \resizebox{\textwidth}{!}{
    \begin{tabular}{|l|c|c|c|c|c|}
        \hline
        Data & $\theta$  &  $f_{\theta}(t)$  & $\ig(f_\theta, t)$ & $\Delta \theta_r$ & $\ig(f_\theta, t=96)$\\ \hline
        Step   & $a$     &  $a$              & $\Delta a_r$ & 0.5 & 0.5   \\ 
        Ramp   & $(a,b)$ &  $a + bt$         & $(\Delta a_r, \Delta b_r t)$ & (-6, 0.1) & (-6, 12) \\ 
        SinACP   & $(\alpha, \phi)$  & $\alpha \cos(2\pi t + \phi)$  & $(\Delta \alpha_r \Delta g_r(t), \Delta f_r (t) - \Delta \alpha_r \Delta g_r (t))$ & (0.2, 0.01) & (0.2, 0.004) \\ \hline
    \end{tabular}
    }
    \label{tab:experiment}
\end{table}

\subsection{Frequency change point}

\begin{figure}
  \centering
  % First row
  \begin{subfigure}[b]{0.4\textwidth}
    \centering
    \includegraphics[width=\textwidth]{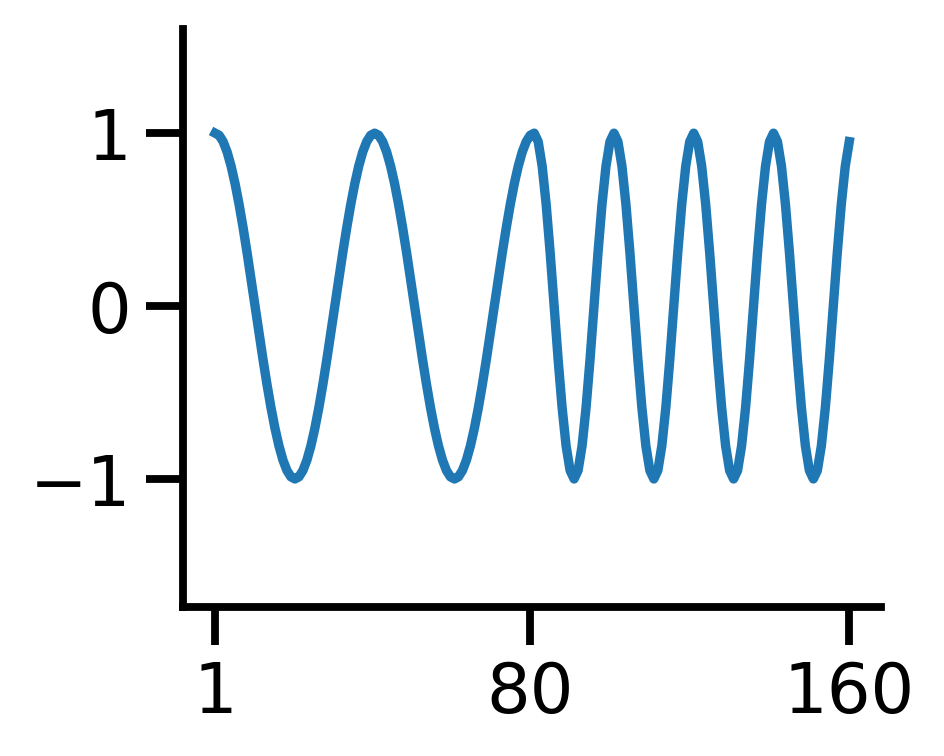}
    \caption{}
    \label{fig:fm1}
  \end{subfigure}%
  \begin{subfigure}[b]{0.4\textwidth}
    \centering
    \includegraphics[width=\textwidth]{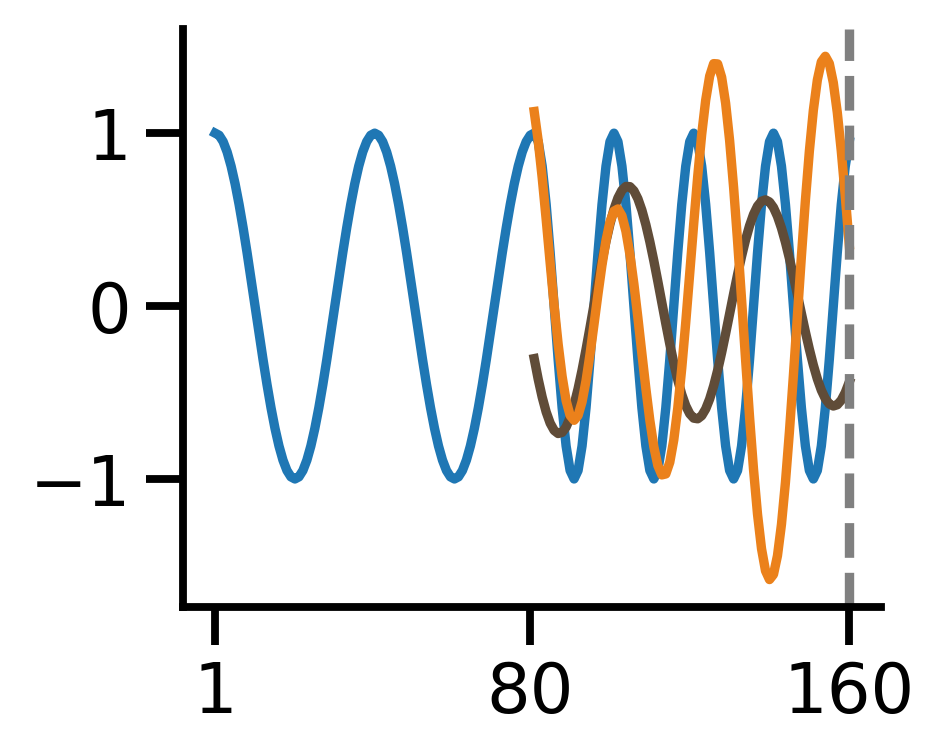}
    \caption{}
    \label{fig:fm2}
  \end{subfigure}%
  \begin{subfigure}[b]{0.15\textwidth}
    \centering
    \raisebox{36pt}{\includegraphics[width=\textwidth]{legend_2.png}}
    %\caption{}
    \label{fig:fm3}
  \end{subfigure}%
  \caption{Experimental evaluation on synthetic time-series with frequency and phase change point.}
  \label{fig:synthetic2}
\end{figure}

We now apply the proposed method to the more challenging case of a sinusoidal time series of length $n=160$, featuring a frequency and phase change point at $t=81$; see Figure \ref{fig:fm1}. This time series corresponds to discrete observations derived from a linear ODE with time-dependent coefficients; see Appendix \ref{sec:ode}. The BAPC is applied taking a correction window size of $r=80$ and the 1-Nearest-Neighbor correction model. We use the AR formulation \eqref{eq:ar} as the base model, where the parameters $\bvarphi= (\varphi_1,\varphi_2)$ are estimated via {\em robust} auto-regression. 

Robust AR on a time series $(z_t)_{t=1}^n$ involves first removing outliers from the set $\{x_t\}_{t=3}^{n} := \{(z_{t-1}, z_{t-2})\}_{t=3}^{n}$ and then performing linear regression of $\{z_t\}_{t \in I}$ against $\{x_t\}_{t \in I}$, where $I$ is the set of indices after outlier removal. The time series in Figure \ref{fig:fm1} exhibits no discontinuities; hence, no outliers are removed during the first application of the robust AR base model. On the other hand, applying robust AR after correction, removes the outlier $x_{82} = (y_{81}^\prime, y_{80}^\prime) = (y_{81} - \hat{\varepsilon}_{81}, y_{80})$, thereby alleviating the model misspecification induced by the discontinuity in $y^\prime$ at $t = 81$.  Figure \ref{fig:fm2} shows the base model after correction (brown line $f_{\theta_r}$) and the surrogate model (orange line $f_{\theta_0} + \Delta f_r$). 

The first row in Table \ref{tab:synthetic2} shows the BAPC configuration and results at $t=160$, where, in particular, the explanation and integrated gradient were computed from the formula at Proposition \ref{prop:ig_formulas}-3. Interpreting these results is not straightforward; hence, we make use of Proposition \ref{prop:ar2sin} to analyze the BAPC results in the context of the sinusoidal base model $f_\theta(t) = \alpha \exp(-\beta t) \cos(\omega t + \varphi)$, with $\theta = (\alpha, \beta, \omega, \varphi)$. From Proposition \ref{prop:ig_formulas}-2, we obtain the integrated gradient $\ig(f_\theta, t=160)$ value shown in the second row of Table \ref{tab:synthetic2}. This value identifies a significant increase in frequency and a minor adjustment in phase, both of which are present in the input data and captured by the 1-Nearest-Neighbor correction. Additionally, it indicates a small change in the exponential decay parameter, which is absent in the data. Such inaccuracy could potentially be mitigated by increasing the sampling frequency, at the expense of the high floating-point precision required to compute the combinatorial formulas \eqref{aling:ig1} and \eqref{aling:ig2}.

\begin{table}[ht]
  \centering
  \caption{Experimental configuration and results for sinusoidal data with frequency change point (SinFCP).}
  \resizebox{\textwidth}{!}{
  \begin{tabular}{|c|c|c|c|c|c|}
  \hline
    Data & $\theta$  &  $f_{\theta}(t)$  & $\ig(f_\theta, t)$ & $\Delta \theta_r \times 1\mathrm{e}{2}$ & $\ig(f_\theta, t=160) \times 1\mathrm{e}{2}$\\ \hline
    \multirow{2}{*}{SinFCP} & $(\varphi_1, \varphi_2)$ & Eq. \eqref{eq:ar} & Eq. \eqref{aling:ig1}, \eqref{aling:ig2}  & (-3.1, 0.3) & (46, -6) \\ 
   & $(\alpha, \beta, \omega, \phi)$ & $\alpha\exp(-\beta t) \cos(\omega t + \phi)$ & Eq. \eqref{eq:ig} & (0.4,0.2,6.6,-5.2) & (0.0,-1.9,41.7,	-0.2) \\ 
    \hline
  \end{tabular}
  }
  \label{tab:synthetic2}
\end{table}

While it might seem reasonable to use $f_{(\alpha, \beta, \omega, \phi)} = \alpha \exp(-\beta t) \cos(\omega t + \varphi)$ directly instead of the AR formulation, this approach is not well-suited for the current example. Indeed, it can be shown that $\omega \mapsto \sum_{t=1}^n (y_t - f_{(\alpha, \beta, \omega, \phi)}(t))^2$ has two global minima, attained at the data frequency values before and after the change point. As a result, in this example, this would lead to $\Delta \omega_r = 0$.

Finally, observe that if the base model $f_\theta$ is linear with respect to $\theta$, then the surrogate model belongs to the same model class. More specifically, in this case, $f_{\theta_0} + \Delta f_r  = f_{\theta_0 + \Delta \theta_r}$, as in Figure \ref{fig:syn1} and Figure \ref{fig:syn2}. Note also that $f_{\theta_0} + \Delta f_r \approx f_{\theta_0 + \Delta \theta_r}$ for the scenario depicted in Figure \ref{fig:syn3}, where equality would hold if $\Delta \phi_r = 0$. In general, for the non-linear case, the surrogate model $f_{\theta_0} + \Delta f_r$ does not belong to the same class as the base model, as shown in Figure \ref{fig:fm2}, where $f_{\theta_0} + \Delta f_r$ has non constant amplitude.

\subsection{Trend change point with varying correction window size}\label{sec:windowsize}

\begin{figure}[ht]
  \centering
  \begin{subfigure}[b]{0.25\textwidth}
    \centering
    \includegraphics[width=\linewidth]{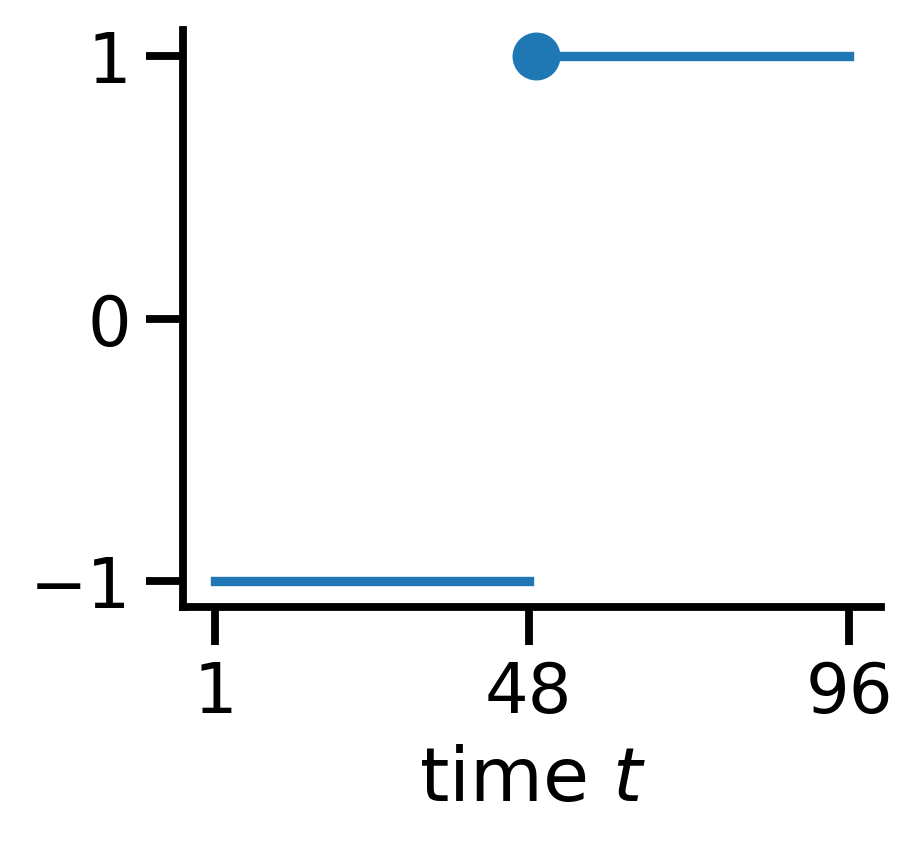}
    \caption{}
    \label{fig:win1}
  \end{subfigure}
  \begin{subfigure}[b]{0.25\textwidth}
    \centering
    \includegraphics[width=\linewidth]{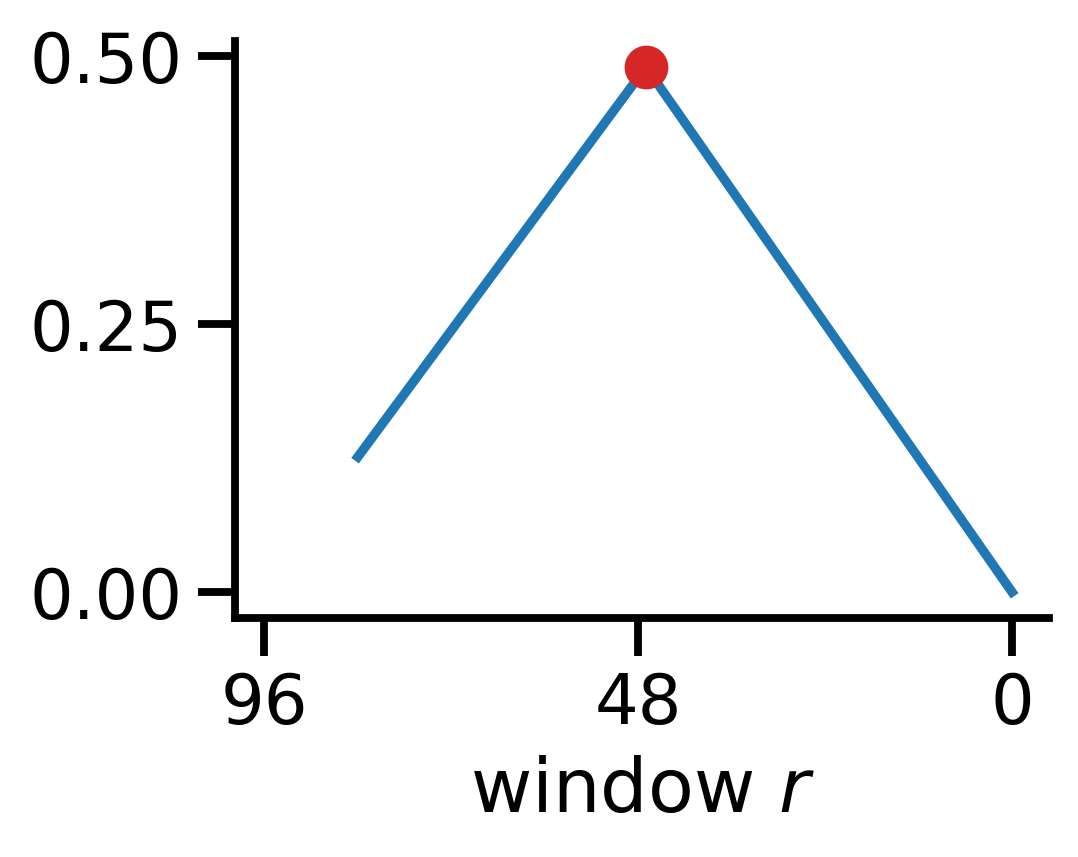}
    \caption{}
    \label{fig:win2}
  \end{subfigure}
  \begin{subfigure}[b]{0.25\textwidth}
    \centering
    \includegraphics[width=\linewidth]{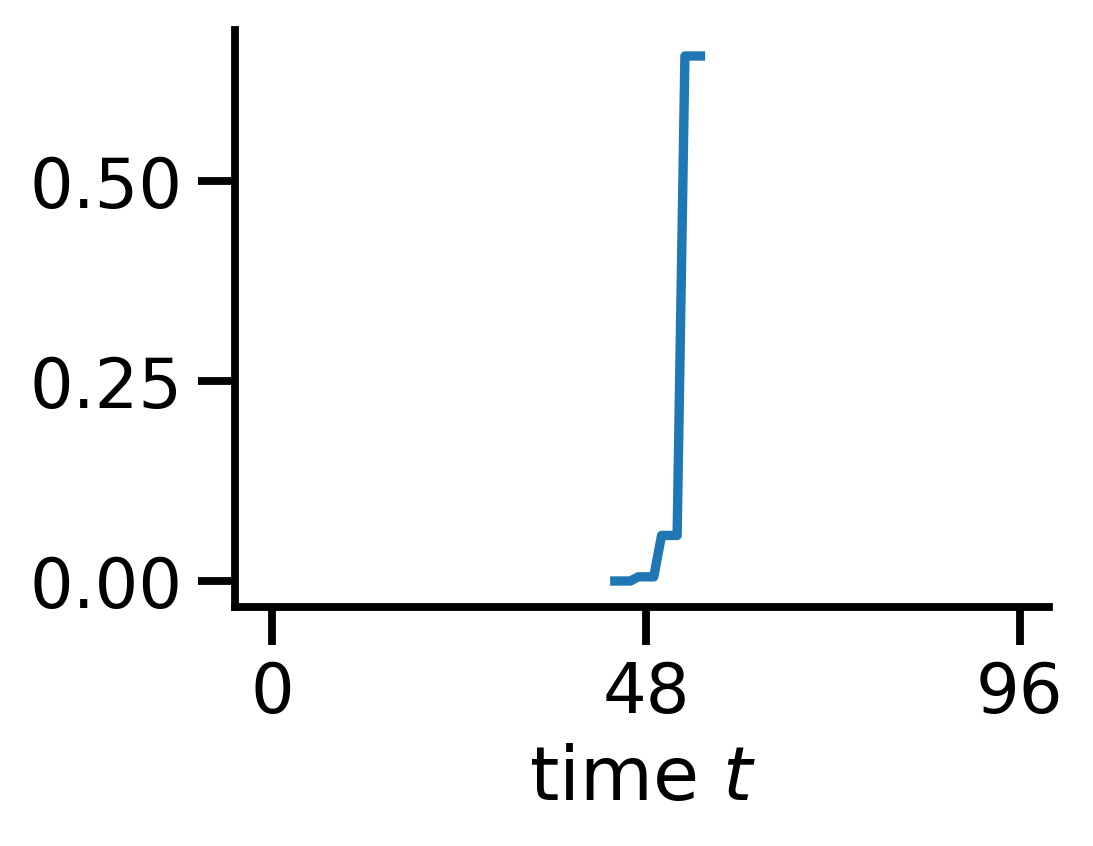}
    \caption{}
    \label{fig:win3}
  \end{subfigure}
  \caption{Experimental evaluation of BAPC for different window sizes over the step input data (a). The surrogate correction  $r \mapsto \Delta f_r(t)$ at $t=96$ (b),  attains a global maximum at  $r=48$. The LIME (c) also provides information on the change point location.}
  \label{fig:winstep}
\end{figure}

\begin{figure}[ht]
  \centering
  \begin{subfigure}[b]{0.25\textwidth}
    \centering
    \includegraphics[width=\linewidth]{2_ramp_data.png}
    \caption{}
    \label{fig:win4}
  \end{subfigure}
  \begin{subfigure}[b]{0.25\textwidth}
    \centering
    \includegraphics[width=\linewidth]{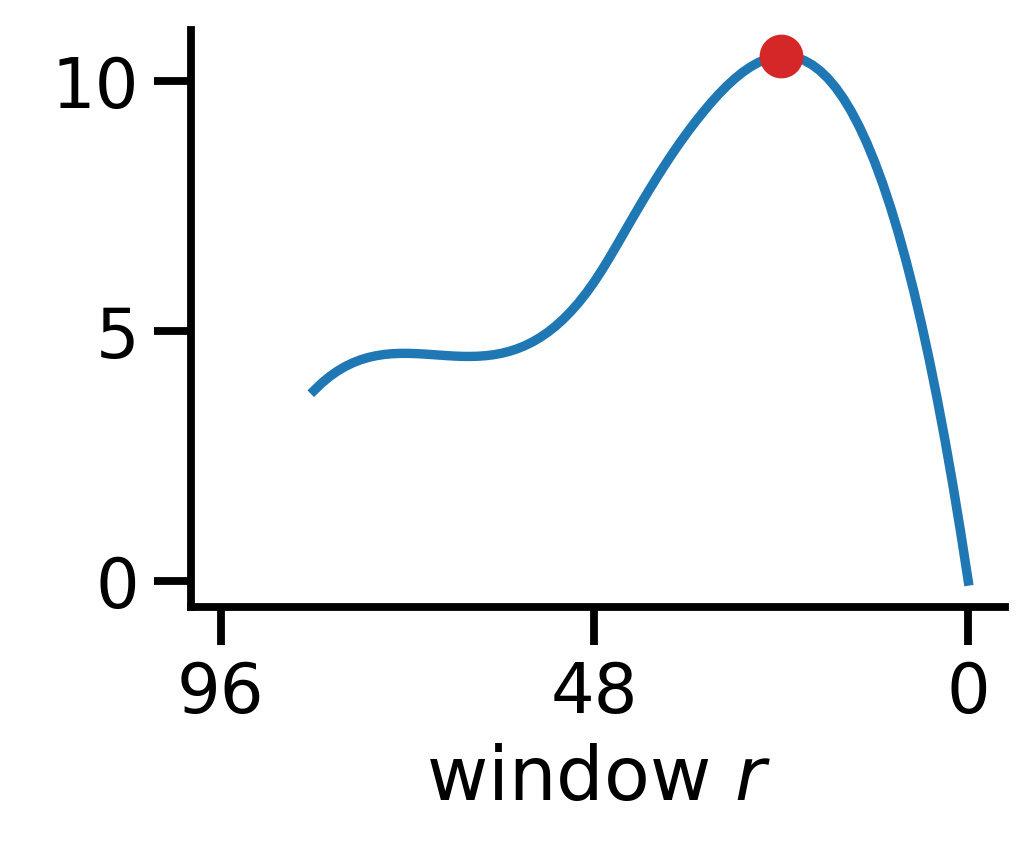}
    \caption{}
    \label{fig:win5}
  \end{subfigure}
  \begin{subfigure}[b]{0.25\textwidth}
    \centering
    \includegraphics[width=\linewidth]{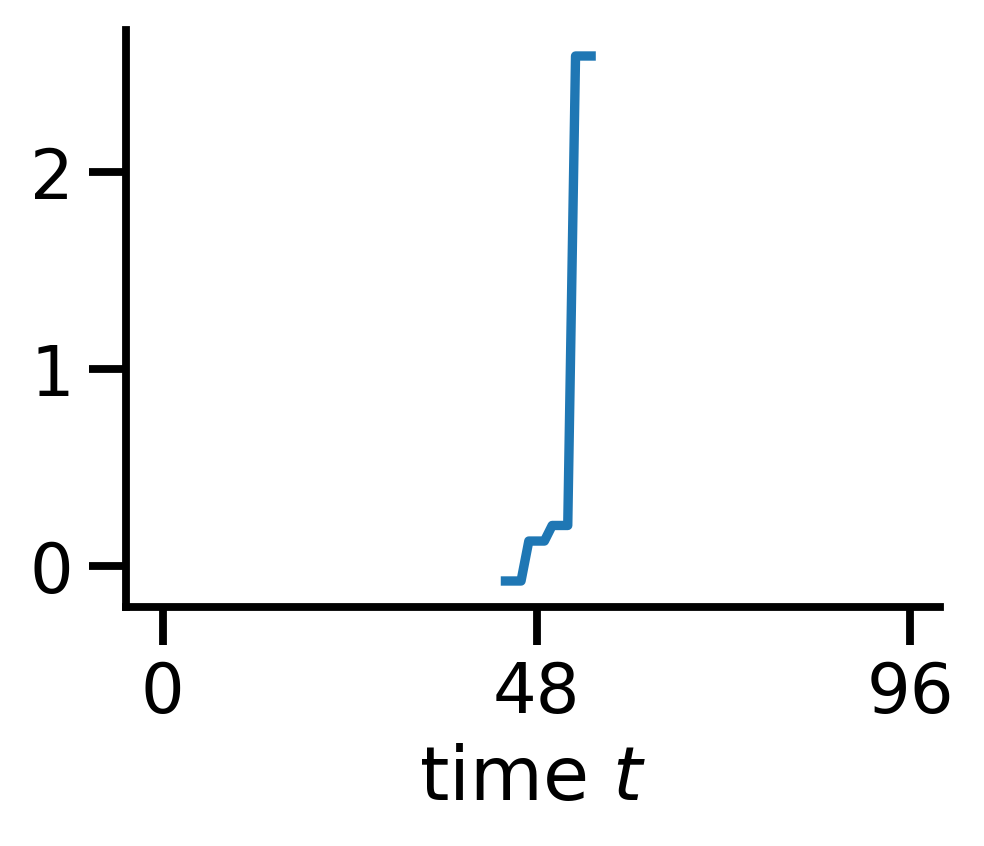}
    \caption{}
    \label{fig:win6}
  \end{subfigure}
  \caption{Experimental evaluation of BAPC for different window sizes over the ramp input data (a). The surrogate correction  $r \mapsto \Delta f_r(t)$ at $t=96$ (b),  attains a global maximum at  $r=24$. The LIME (c) also provides information on the change point location.}
  \label{fig:winramp}
\end{figure}

In this section, we investigate the impact of the correction window size on the BAPC and provide insights into its optimal selection. The step and ramp time series introduced in Section \ref{sec:tren}, depicted in figures \ref{fig:win1} and \ref{fig:win4}, respectively, are used as input data. For this experiment, we employ the base models specified in Table \ref{tab:experiment}, with the correction model implemented as an LSTM recurrent neural network \cite{HochSchm97}. This widely used time-series forecasting model is known for its ability to capture complex data patterns without requiring explicit assumptions about the underlying system's mathematics or physics. We choose an auto-regressive order $p=12$ which provides a trade-off between lagged dependency and the train set size at each time step $s$. We then compute the BAPC by varying the window size across a range of values, $r = 0, \ldots, n$.

To determine the optimal window size at a given time $t$, we propose using $r \mapsto \abs{\Delta f_r^s(t)}$ as the objective function. Using this criterion, the optimal window size is found to be $r = 48$ and $r = 24$ for the step and ramp input data, respectively, as illustrated in figures \ref{fig:win2} and  \ref{fig:win5}. Our results suggest that the optimal window size depends on the type of change-point or anomaly in the data, as well as the AI-correction framework and the choice of the base model. Future research will further explore this dependency and refine the proposed approach.

It is also of interest to compare our approach to other explainable time-series methods. LIME \cite{LIME} is an explainable AI framework designed to provide locally interpretable explanations for complex machine learning models. In the context of time series and auto-regressive models, LIME can be used to understand the impact of individual time steps or features on the model's predictions \cite{tslime}. The process involves perturbing the input data, generating predictions, and fitting a local interpretable model.  

Let us briefly describe the application of LIME in the context of this paper. For a residual time series $(\varepsilon_t)_{t=1}^n$ and a (previously fit) correction model $\widehat{\varepsilon}$ of auto-regressive order $p$, we define $\{x_t\}_{t=p+1}^{n} := \{(\varepsilon_{t-1}, \ldots, \varepsilon_{t-p})\}_{t=p}^{n}$. Given an instance $x_t$, the goal of LIME is to explain the black-box prediction $\widehat{\varepsilon}(x_t)$ in terms of individual coordinates of $x_t$. To achieve this, a partition of $x_t$ consisting of contiguous segments of suitable sizes, is computed. A new set of $k$ perturbed samples $x_1^\prime, \ldots, x_k^\prime$ is then generated by randomly selecting segments of the partition and replacing the values in these segments with a non-informative placeholder. Next, a linear regression of the predictions $y_1^\prime, \ldots, y_k^\prime$ against the perturbed samples $x_1^\prime, \ldots, x_k^\prime$ is performed, where $y_1^\prime := \widehat{\varepsilon}(x_1^\prime), \ldots, y_k^\prime := \widehat{\varepsilon}(x_k^\prime)$.  Finally, the resulting vector $c_t$ of $p$ coefficients from the regression is interpreted as a measure of how each input data point $(\varepsilon_{t-1}, \ldots, \varepsilon_{t-p})$ contributes to the prediction $\widehat{\varepsilon}_{t} := \widehat{\varepsilon}
(\varepsilon_{t-1}, \ldots, \varepsilon_{t-p})$.

We applied LIME to the residuals and the correction model obtained after Step-2 of the BAPC, resulting in coefficients $c_t = (c_{t1}, \ldots, c_{tp})$ for $t = p+1, \ldots, n$. Figures \ref{fig:win3} and \ref{fig:win6} display $c_t$ plotted against $t-p, \ldots, t-1$ for $t = 56$, corresponding to the step and ramp data, respectively. The time point $t = 56$ was chosen for visualization due to its proximity to the change point. A change in the coefficient patterns associated with the change point is observed, aligning with the findings from BAPC. We suggest that BAPC and LIME have the potential to provide valuable complementary insights, subject to further adaptations.

\subsection{Trend change point on sequential setting}\label{sec:windowloc}

\begin{figure}[ht]
  \centering
  \begin{subfigure}{0.45\textwidth}
    \centering
    \includegraphics[width=\linewidth]{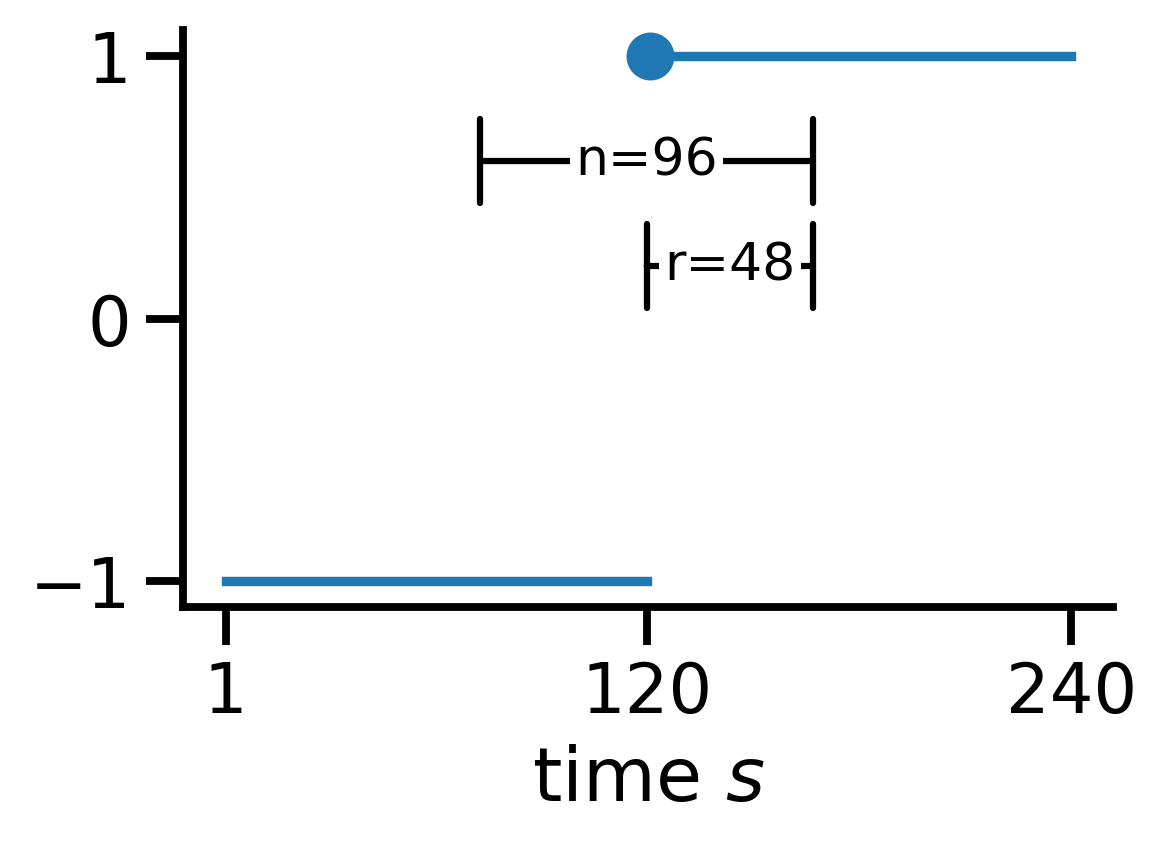}
    \caption{}
    \label{fig:seq1}
  \end{subfigure}
  \begin{subfigure}{0.45\textwidth}
    \centering
    \includegraphics[width=\linewidth]{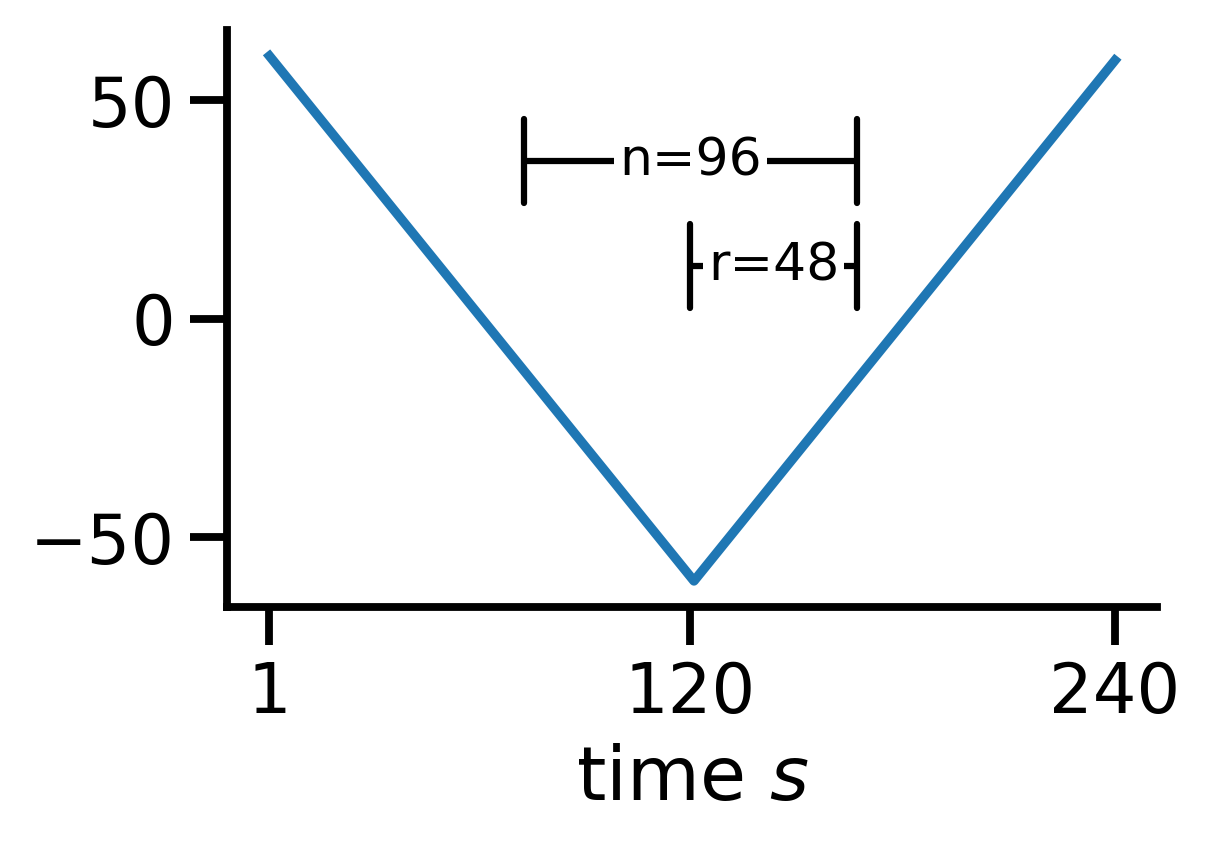}
    \caption{}
    \label{fig:seq2}
  \end{subfigure}
  \caption{Input step (a) and ramp (b) time series for the sequential BAPC. The horizontal bracketed lines depict the sliding window centered at the change point.}
  \label{fig:seqdata}
\end{figure}

Let us now apply the sequential BAPC to the step time series $(y_s)_{s=1}^m$ of length $m = 240$, shown in Figure \ref{fig:seq1}. For each $s = n, \ldots, m$, we apply BAPC to the segment $(y_t)_{t=s-n+1}^{s}$ with $n = 96$, $r = 48$, and the base model $f_{\theta}(t) = \alpha$,  where $\theta = \alpha$. For the correction model, we utilize an LSTM recurrent neural network with auto-regressive order $p=12$. We also applied the sequential BAPC to the ramp data shown in Figure \ref{fig:seq2} under the same setting described before, but taking the base model base model $f_{\theta}(t) = \alpha + \beta t$,  where $\theta = (\alpha,\beta)$.

\begin{figure}[ht]
  \centering
  % First row
  \begin{subfigure}{0.44\textwidth}
    \centering
    \includegraphics[width=\textwidth]{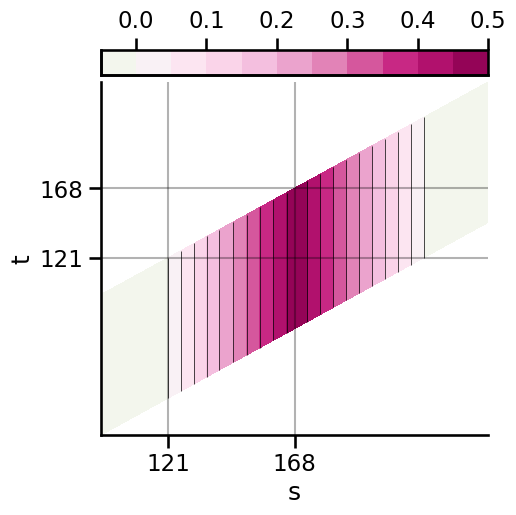}
    \caption{}
    \label{fig:seq3}
  \end{subfigure}
  \begin{subfigure}{0.44\textwidth}
    \centering
    \includegraphics[width=\textwidth]{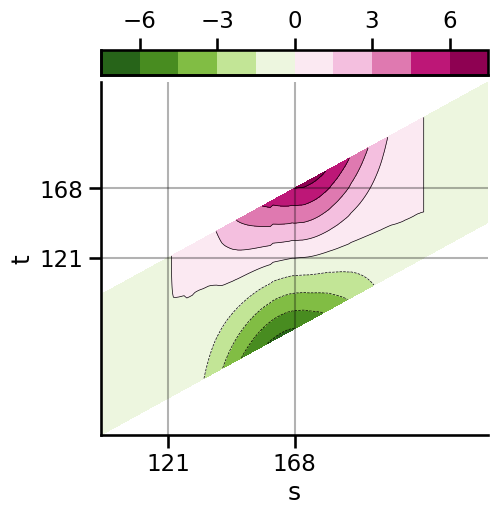}
    \caption{}
    \label{fig:seq4}
  \end{subfigure}

  \caption{Sequential BAPC results: For the step (a) and ramp (b) data, the surrogate correction $(s, t) \mapsto \Delta f_r^s(t)$ attains (non-unique) extreme values when $s = 168$, which corresponds to the training window of size $n$ being centered at the change point; see Figure \ref{fig:seqdata}.}
  \label{fig:synthetic4}
\end{figure}

Observe that from Proposition \ref{prop:completeness}, we have  
\begin{equation}
    \sum_{k=1}^{q} \ig_k(f_{\theta}, s, t) = \Delta f_r^s(t),
\end{equation}  
which justify the use of $(s, t) \mapsto \Delta f_r^s(t)$ as an overall score to assess changes in the BAPC outcome over time. In this context, we aim to investigate the local maxima and minima of the mapping $(s, t) \mapsto \Delta f_r^s(t)$ to understand the behavior of the underlying AI correction. To this end, let us examine the heatmaps of $\Delta f_r^s(t)$ for $s = n, \ldots, m$ and $t = s-n+1, \ldots, s$, displayed in Figure \ref{fig:synthetic4} for the case of the (a) step and (b) ramp data. 

From Figure \ref{fig:seq3}, we observe that for the step data case, the surrogate correction $(s, t) \mapsto \Delta f_r^s(t)$ attains a global maximum at $s = 168$ and for every $t$ in the associated training window, $t = s-n, \ldots, s$. For the ramp data case, shown in Figure \ref{fig:seq4}, the surrogate correction $(s, t) \mapsto \Delta f_r^s(t)$ reaches a global minimum at the left endpoint of the correction window ($t = s-n$) and a global maximum at the right endpoint ($t = s$). The situation $s = 168$ corresponds to the training window being centered at the change point, as depicted in Figure \ref{fig:seqdata}. These observations suggest that the local maxima and minima of $(s, t) \mapsto \Delta f_r^s(t)$ are associated with change points in the data, provided they are captured by the AI corrections. A detailed investigation of these optimal points and their relationship with $n$ and $r$ lies beyond the scope of this paper and will be addressed in future research. 

\subsection{Air passenger dataset}

\begin{figure}[ht]
  \centering
  \includegraphics[width=0.9\linewidth]{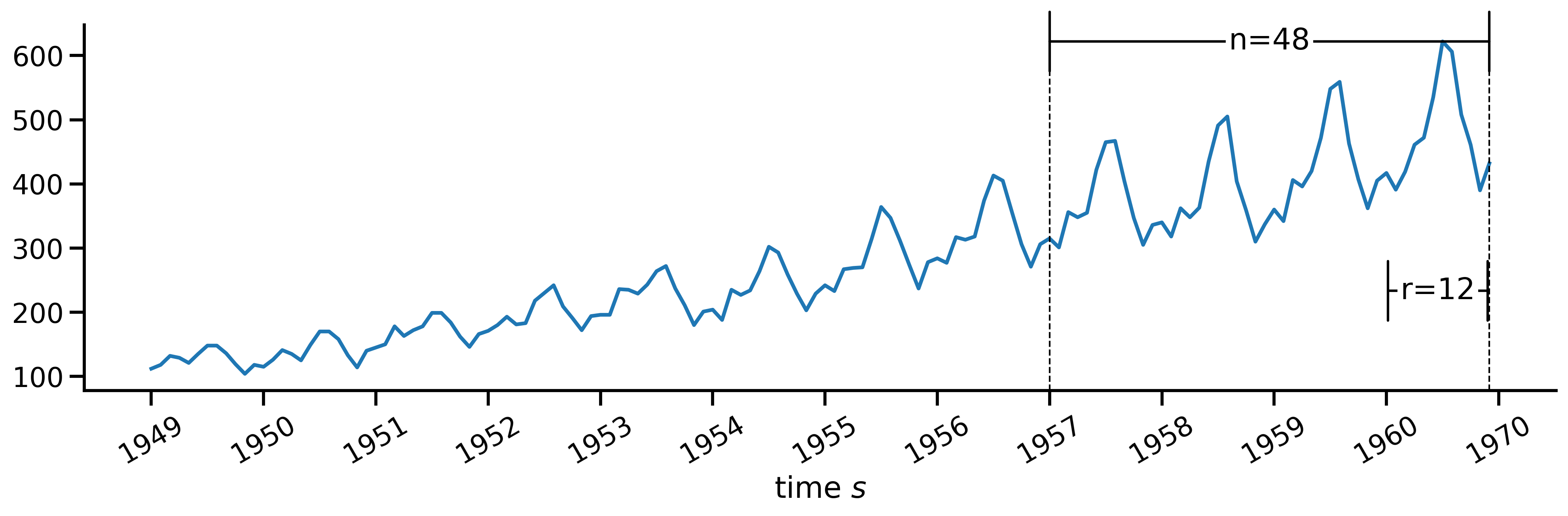}
  \caption{Air passenger time series.}
  \label{fig:airtimeseries}
\end{figure}

We continue the experimental validation by applying the proposed method to the air-passenger dataset, a well-known dataset commonly used in time series analysis and machine learning studies. This dataset contains $m = 144$ monthly totals of international airline passengers (in thousands) from January 1949 to December 1960; see Figure \ref{fig:airtimeseries}.  For this experiment, we aim to analyze how passenger demand evolves over time on an annual basis, with a specific focus on the last four years of the dataset. 

To achieve this, we compute the sequential BAPC with a training window size of $n = 48$, a correction window size of $r = 12$, base model 
\begin{equation}
    f_{\theta}(t) = a + bt + ct^2 + \alpha \cos\left(\frac{2\pi}{12} t + \phi\right), 
\end{equation}
where $\theta = (a, b, c, \alpha, \phi)$ represents the intercept, slope, quadratic term, amplitude, and phase. This model combines a polynomial trend with a periodic pattern. The LSTM recurrent neural network of auto-regressive order $p=12$ is used as the correction model. 

\begin{figure}[ht]
  \centering
  \begin{subfigure}{0.3\textwidth}
    \includegraphics[width=\linewidth]{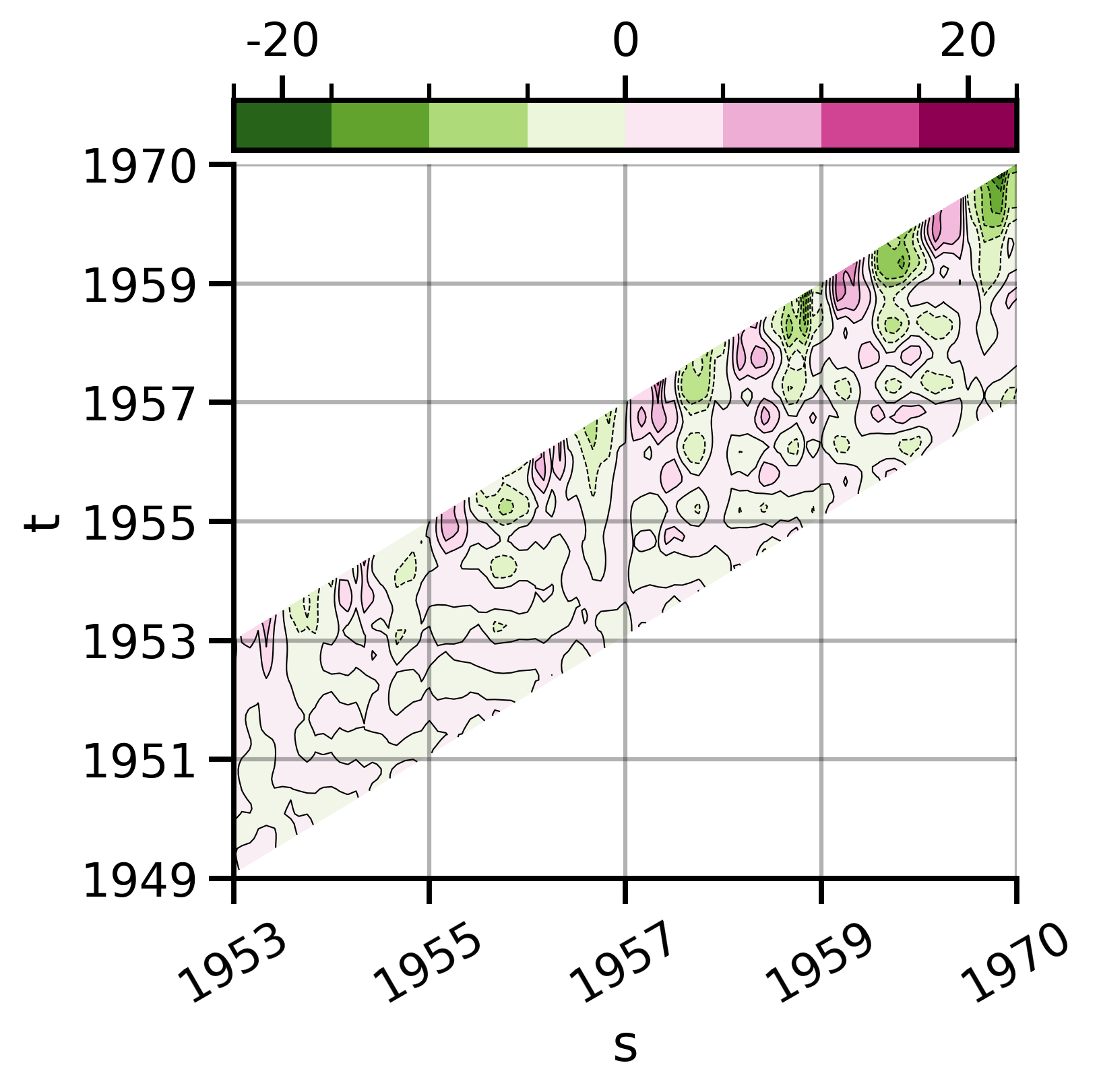}
    \caption{Surrogate correction}
    \label{fig:air1}
  \end{subfigure}
  \begin{subfigure}{0.3\textwidth}
    \includegraphics[width=\linewidth]{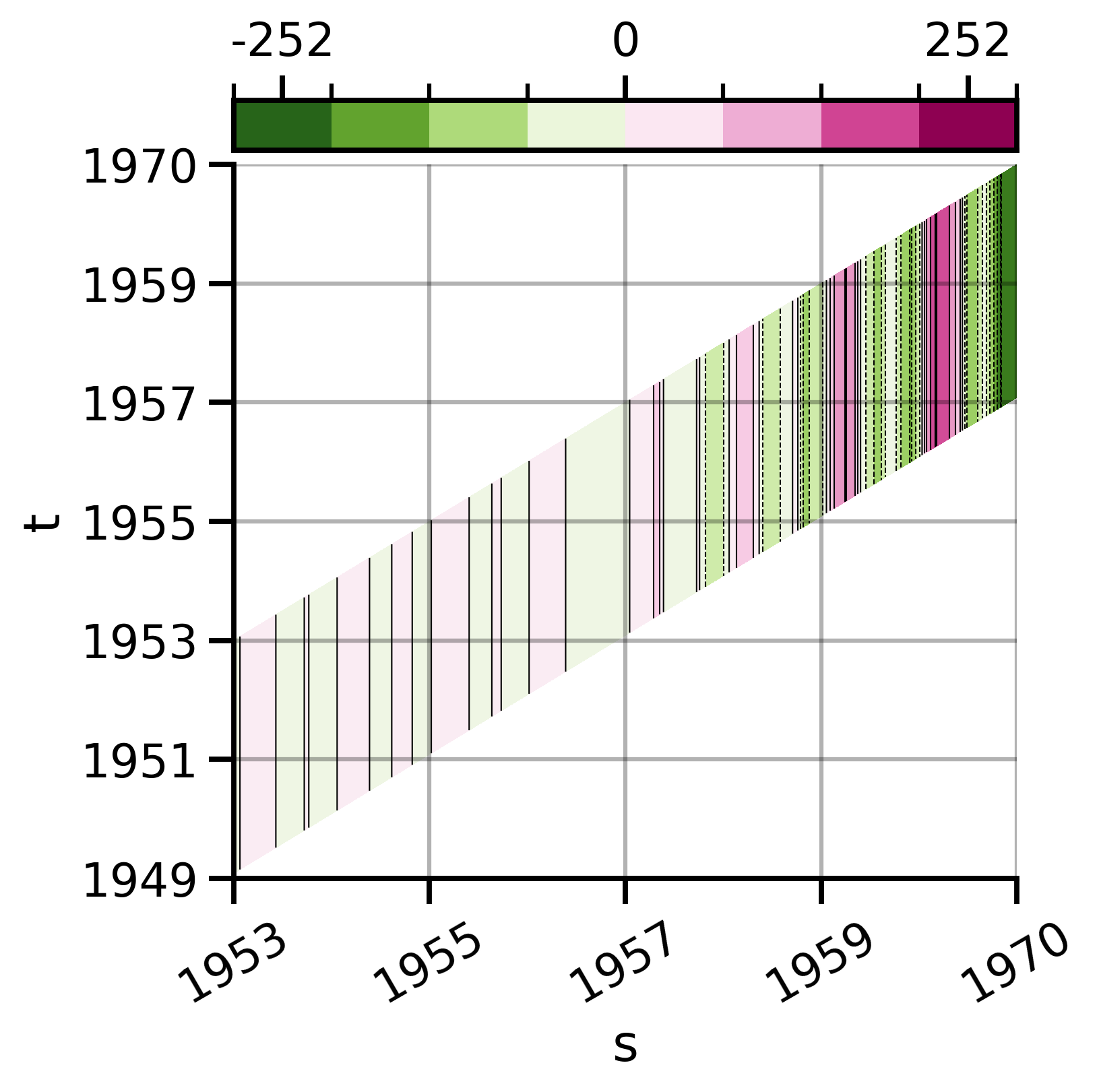}
    \caption{Intercept}
    \label{fig:air2}
  \end{subfigure}
  \begin{subfigure}{0.3\textwidth}
    \includegraphics[width=\linewidth]{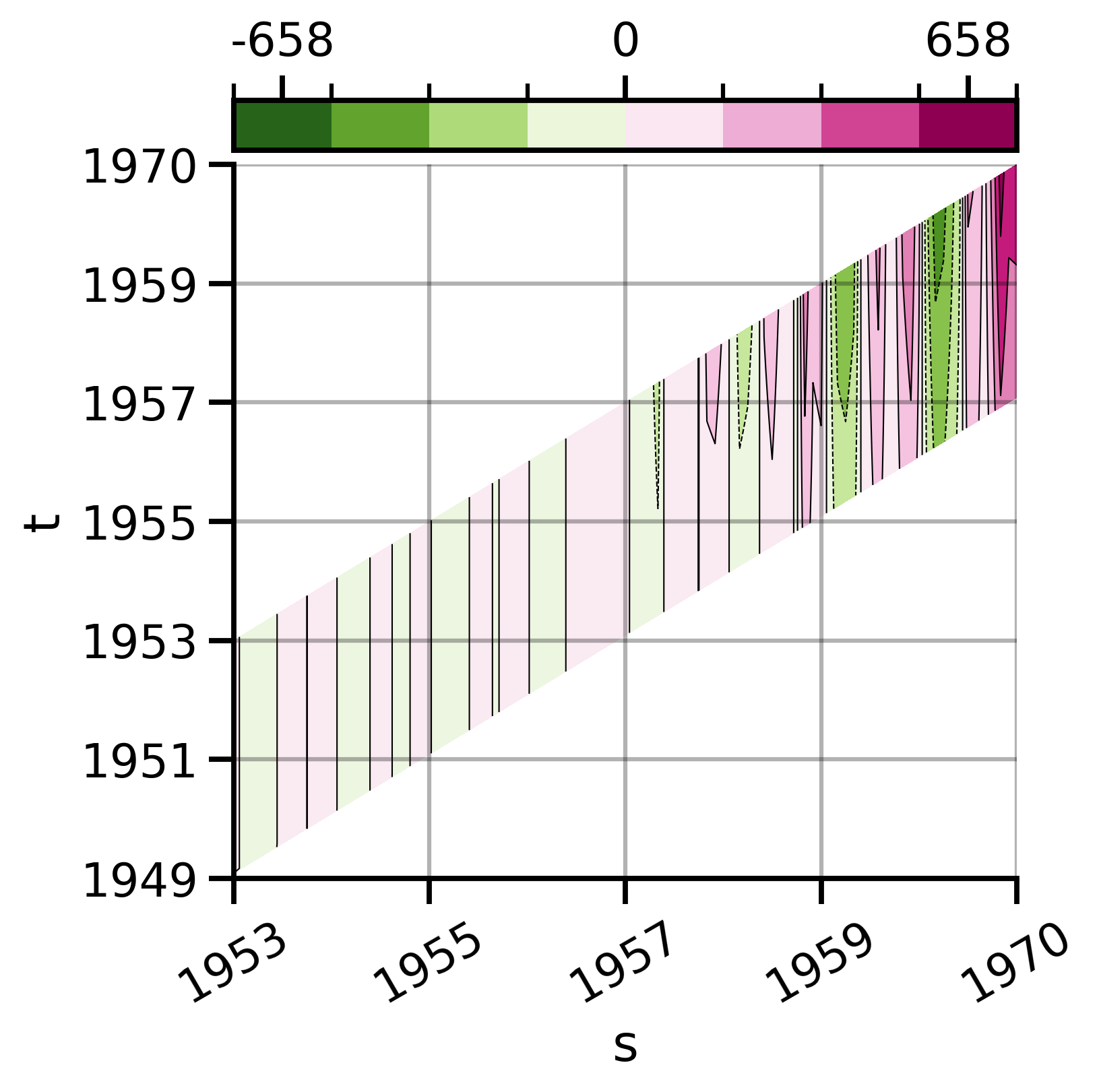}
    \caption{Slope}
    \label{fig:air3}
  \end{subfigure}
  \par\bigskip
  \begin{subfigure}{0.3\textwidth}
    \includegraphics[width=\linewidth]{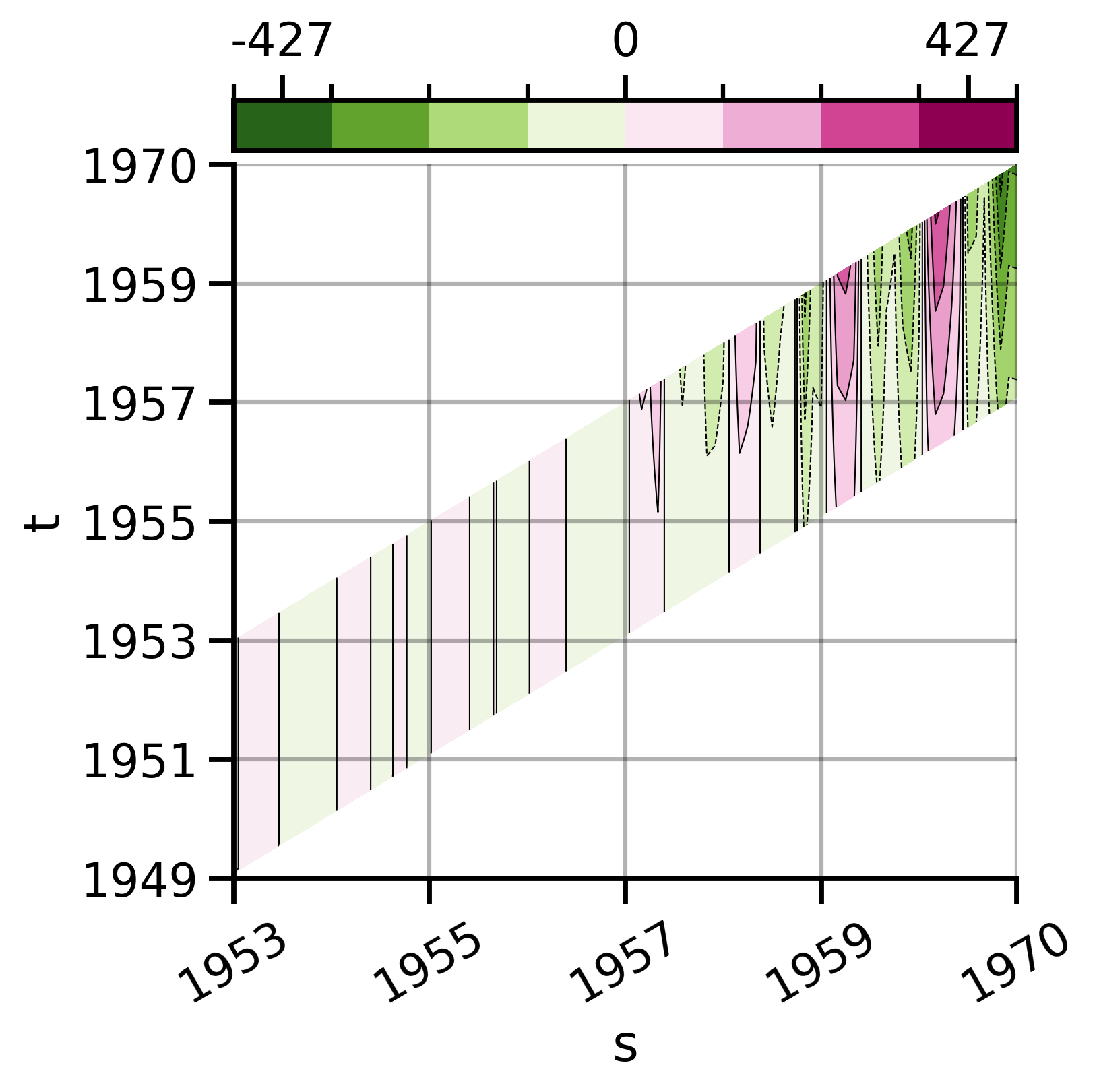}
    \caption{Quadratic}
    \label{fig:air4}
  \end{subfigure}
  \begin{subfigure}{0.3\textwidth}
    \includegraphics[width=\linewidth]{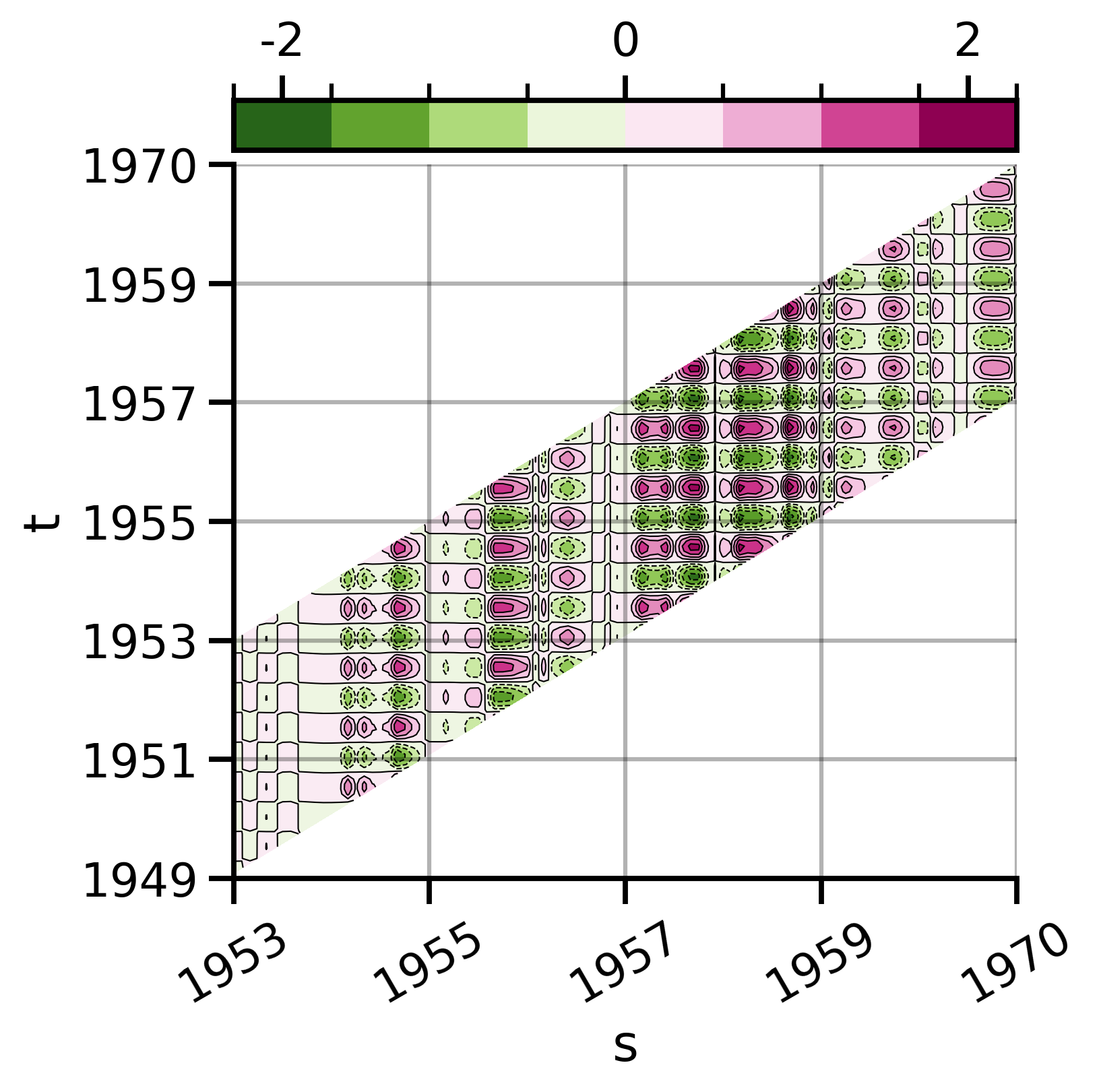}
    \caption{Amplitude}
    \label{fig:air5}
  \end{subfigure}
  \begin{subfigure}{0.3\textwidth}
    \includegraphics[width=\linewidth]{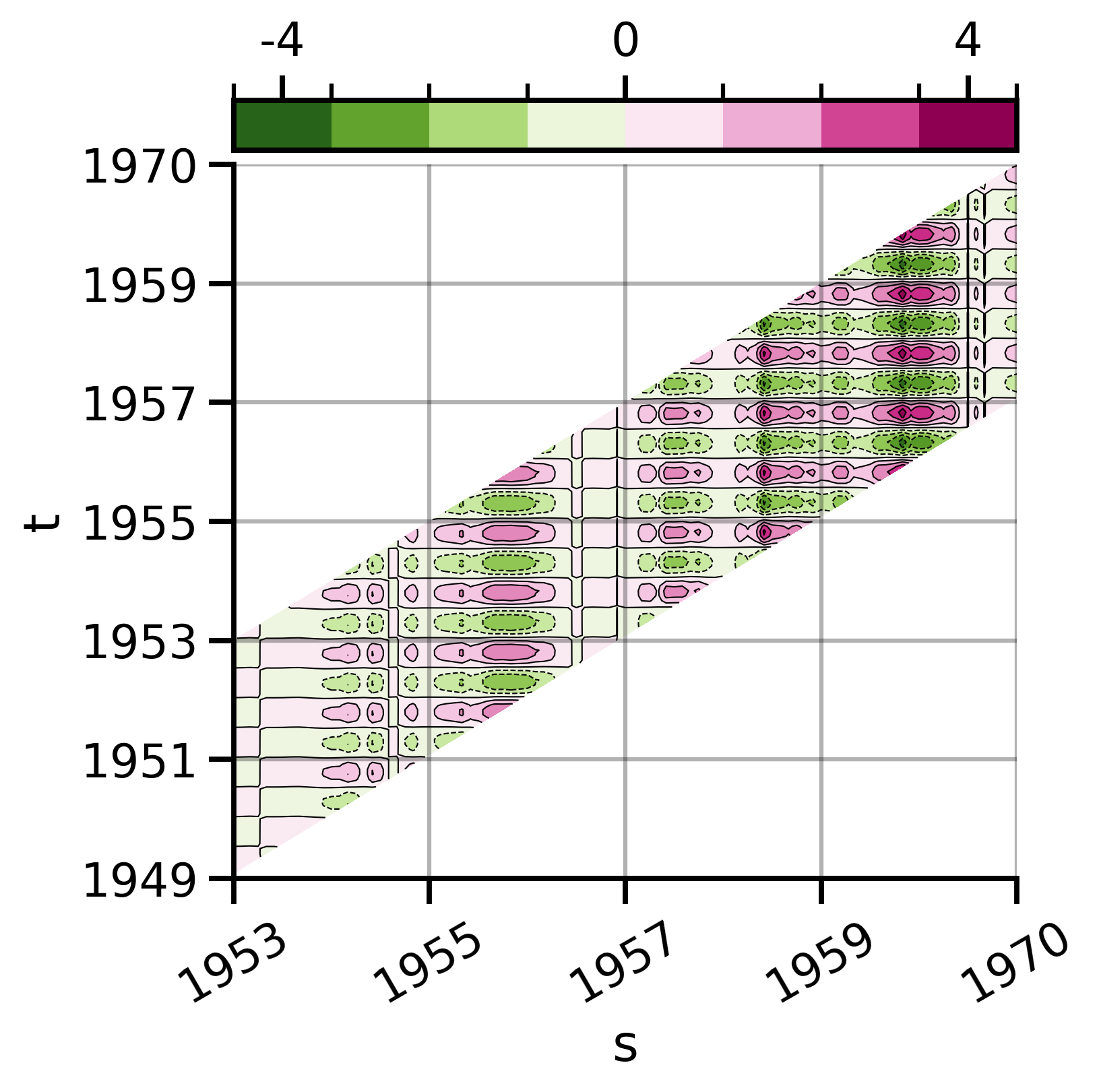}
    \caption{Phase}
    \label{fig:air6}
  \end{subfigure}
  \caption{Sequential BAPC on air-passenger data. The heatmaps corresponds to (a) surrogate correction $(s, t) \mapsto \Delta f_r^s(t)$ and the integrated gradient $(s, t) \mapsto \ig_k(f_{\theta}, s, t)$ associated to the intercept $a$, slope $b$, quadratic term $c$, amplitude $\alpha$, and phase $\phi$: The strong coloring show the intervals of the features being particularly significant for the correction $\widehat{\epsilon}$ (e.g., the importance of an upward directed quadratic term after 1959, well visible in Fig. 9). }
  \label{fig:airdata}
\end{figure}

The surrogate correction $(s, t) \mapsto \Delta f_r^s(t)$ is shown in Figure \ref{fig:air1}. From the figure, we observe that the surrogate correction attains its minimum and maximum values for $s = 1960$. In other words, there is not only a noticeable increase in the general trend of the data, as apparent from Figure \ref{fig:airdata}, but also an increase in the complexity of the corrections applied on top of that trend, as modeled by the black-box LSTM. The surrogate correction quantifies the magnitude of the AI correction.

To enhance explainability, we utilize Proposition \ref{prop:completeness} and Proposition \ref{prop:ig_formulas} to decompose the surrogate correction $(s, t) \mapsto \Delta f_r^s(t)$ (Figure \ref{fig:air1}) into its integrated gradient components $(s, t) \mapsto \ig_k(f_{\theta}, s, t)$ associated with the base model parameters: the intercept $a$, slope $b$, quadratic term $c$, amplitude $\alpha$, and phase $\phi$. From these figures, we observe that most of the surrogate correction is attributed to the polynomial trend. However, it is also noteworthy that the amplitude integrated gradient reveals interesting patterns, while the phase integrated gradient suggests that the corrections maintain a degree of continuity.

In conclusion, the BAPC framework delivers explainability for the black-box correction model (in this case, the LSTM) by framing it within the structure of the chosen base model. We believe that, in applications requiring explainability, additional expert knowledge is crucial for configuring the method (e.g., selecting the training window, correction window, base model, and correction model) and interpreting the results of the proposed approach.

As is noted in \cite{atm}, a review on XAI methods in air traffic control, "Indeed, developed XAI systems are more directed to the developer or the debugger than the final end-user" \cite{mueller}. In contrast, the new plot-type used in Figures 8 and 10 may be considered a diagram type particularly useful for aplications with demand for fast, easy-to-understand {\em online} XAI methods.

\subsection{Tennessee Eastman Process}
\begin{figure}[ht]
	\centerline{\includegraphics[width=1\textwidth]{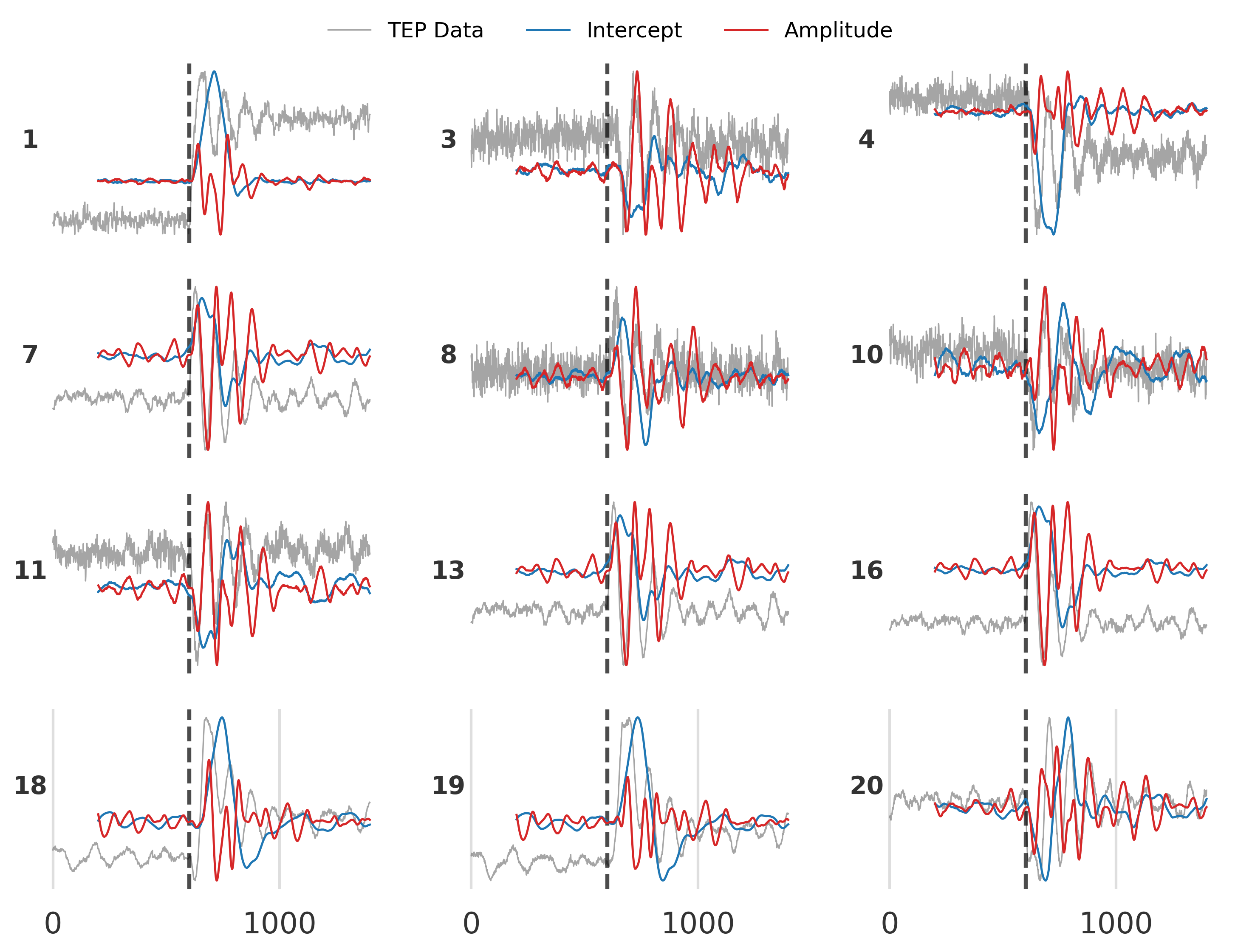}}
	\caption{Sequential BAPC on TEP data: Comparison of time-dependent relative feature importance.}
	\label{fig:tep}
\end{figure}

We conclude the experimental evaluation by applying our method to the Tennessee Eastman Process (TEP) dataset, a widely used simulation-based benchmark for evaluating process monitoring techniques~\cite{MELO2022107964}. The TEP dataset simulates multivariate time series from the control and monitoring of a chemical plant under various process disturbances. For our evaluation, we selected 12 continuous process variables (out of the 22 available), covering a diverse mix of flow rates, temperatures, and pressures, making them well-suited for capturing fault-induced dynamic changes. We used the Python package \texttt{BibMon}~\cite{MELO2024100182} to load and manage the dataset.
We focused on fault scenario 1, which corresponds to a step change in the A/C feed ratio. This fault simulates a sudden shift in the proportion of two key reactants entering the process, thereby disrupting the stoichiometry of the chemical reaction. The choice of this fault is motivated by its significant and immediate impact on process stability, particularly affecting upstream variables such as the flow rate, pressure, and temperature. Its dynamics are well-suited for testing our XAI method, as it introduces observable deviations, including step changes and oscillatory behavior, in several of the selected variables. Figure~\ref{fig:tep} displays the selected time-series variables (shown in gray, with their variable IDs listed on the left-hand side). Each displayed time series has 1400 observations.

We compute the sequential BAPC with a training window size of $n = 200$, a correction window size of $r = 100$, and a base model
\begin{equation}
    f_{\theta}(t) = a + \alpha \cos\left(\frac{2\pi}{100} t + 0.4 \pi \right),
\end{equation}
where $\theta = (a, \alpha)$ represents the intercept and amplitude, combining the step and oscillatory patterns. For the correction, we choose the XGBoost model, a well-known gradient boosting technique based on decision trees, that offers a good trade off between accuracy and computational efficiency, making it well suited for this final large-scale evaluation \cite{ChenGuestrin2016}.
The integrated gradient $s \mapsto \ig_k(f_{\theta}, s, s)$ (See \eqref{eq:igst}) for the intercept $(k = 1)$ and amplitude $(k = 2)$ are shown in Figure~\ref{fig:tep} in blue and red, respectively. As noted in Observation~\ref{obs:ig}, $\ig_k(f_{\theta}, s, s)$ (with $s$ in both temporal arguments) corresponds to the importance of the $k$-th parameter for the prediction at the current time. Therefore, the mapping $s \mapsto (\ig_1(f_{\theta}, s, s), \ig_2(f_{\theta}, s, s))$ can be interpreted as a dynamic feature-importance ranking that explains the black-box correction within the current correction window. 
For example, time series 1 in Figure~\ref{fig:tep} exhibits an apparent step and oscillatory change point at the fault time, which is recovered by the integrated gradient curves, thus providing a meaningful explanation. We also observe that, for this case, the integrated gradient curves return to close to zero once the time series stabilizes.

Eminence of the features "Intercept" and "Amplitude" from the base model to explain the correction around the anomaly are monitored within the correction window which is passing across it in a sliding fashion. 

These plots (and those in Figs. 8-10) may reasonably be compared to ICE-plots \cite{ice} which also show the  feature importance in connection with the change of the variable relative to specific instances, the index of which, however, is played by the time-index, which provides us with a natural choice of other 'close' instances. Other than ICE plots, not the whole range of the input feature's possible values is systematically explored. Instead a range of patterns occurring on the given data stretch of the time series is 'scanned'. While the type of plot in Fig. 11 allow direct relative comparison of features at every instance, the plots in Fig. 8-10 allow the analysis of single features eminence over time.

\section{Conclusions}\label{sec:conclusions}

We defined an extension of the BAPC method for time series based on the notion of being able to compare the change  in the model parameters of an interpretable time series model before and after the application of a complex black box correction model. In particular, we defined the concept of sequential BAPC leading to an explainable time series model for each point in time, by means of a time dependent explanation. 

The optimal choice of the correction window  (locality problem in local surrogate modeling; see Question 3 in \cite{LIMErig}) is our primary focus of ongoing research, which we plan to propose in the form of a sequential BAPC with adaptive correction window value. The presented illustrative example provides initial insights, by linking the explanation with the discovery and explanation of change points in the data.

It is seen that BAPC is able to deliver explanations of changes in a sequence of observations by changes of parameters associated with a law of motion if taken as a physical model for observed time series data, linking it to physics informed machine learning \cite{SelMPJ23}. Furthermore, it is able to distinguish the parameters most responsible for this change from others, delivering 'feature-importance' \cite{Carvalho2019} in the sense of local surrogate modeling in explainable AI. Moreover, we also see that knowledge of the underlying physics flow in two directions: initial knowledge of the physics is used in the form of the base model (a sinusoidal function) and then the method delivers back further explanations of the system dynamics (the presence of an amplitude and phase change point). As such the proposed method goes beyond mere data driven explainability, but provides interpretations of predictive models in terms of physical views set up by the choice of the base model. In this context,  we are currently working on extending the results in Section \ref{sec:basemodel} by establishing the connection between auto-regressive models of general order and ODE solutions. Such a framework would enable the use of auto-regressive base models to explain anomalies in terms of the underlying ODE governing the dynamics of the system under study.

\begin{itemize}
    \item \added{SBAPC}\deleted{SPABC} uses the concept from BAPC to break up the action of the full predictive (hybrid) model into an interpretable and a non-interpretable one. \added{SBAPC}\deleted{SPABC} can be given any reasonably chosen base model (e.g., linear regression) modeling the bulk of the phenomena, and allowing for {\bf parameter changes $\Delta f_r$ to explain} the  {\bf correcting AI-model $\widehat{\epsilon}$}. Other surrogate models (such as LIME \cite{LIME}, treeSHAP\cite{treeshap}, or windowSHAP\cite{windowshap}) don't use this decomposition and don't have this freedom of choice. 
    \item The use of the descriptive power of models with physical parameters (such as \eqref{eq:basemodel}) can be used via relations to purely data driven discrete time models (such as Prop. 1-3) for which the definition of SBAPC fits without introducing a discretization error.
    \item SBAPC highlights the feature importance by the help of the integrated gradient concept which helps to break up the corresponding contributions in the specific features cumulative way over the time period of the considered windows. A heatmap (Fig. 10) shows the ranges in which the explanations are strongest (cf. ICE-Plots \cite{ice}). 
    \item The correction window size ($r$) represents the key parameter of the local surrogate approach. Our experiments (see Sect. \ref{sec:windowsize}) show that for typical change points there is a maximum in the strength of the explanation $\Delta f_r$. This coincides with observations about such a maximum made if linear regression is chosen as the base model \cite{BAPC}. 
    \item There is no weighted sub-sampling method as used in LIME. Instead, a specific window size parameter $r$ is singled out which is characteristic of the explanation being strongest (maximal $|\Delta f_r|$). This shows that the BAPC-concept allows an automatic parameter-choice of the local surrogate modeling overcoming LIME's lack of determination \cite{molnar2019} in the size of the neighborhood and therefore delivers a novel solution to the locality problem \cite{locality}. {\bf SBAPC's unique interpretation is that of the most expressive explanation.}
    \item In the BAPC framework, the base model plays a central role by providing an interpretable structure that defines the notion of expected behavior. This model must be specified in advance, based on expert knowledge or hypotheses about the types of patterns or anomalies expected to arise. For instance, if failures are anticipated to exhibit oscillatory dynamics, a sinusoidal base model enables meaningful explanations in terms of frequency or amplitude. Conversely, if step changes dominate, such a model may be ill-suited, resulting in poor interpretability because even changes in step-sizes don't characterize oscillatory behavior. This trade-off becomes particularly salient in online applications, where both timeliness and flexibility are critical. In this way, our approach strongly resembles a current trend in the PINN-approach to XAI (see Section 7.3 of \cite{salience}), in which prior domain knowledge (e.g. in form of a PDE) helps to interpret the AI-model. To investigate this, we include an experiment in which BAPC is deployed in an online setting with both oscillatory and step-like anomalies. The results highlight that while BAPC respects causal constraints and operates sequentially, its explanatory power is contingent on the suitability of the pre-specified base model. This underscores the importance of thoughtful model design or future extensions that adaptively combine multiple candidate structures.
\end{itemize}

Currently, we are working on applying this method to several other industrial use cases and their challenges related to explainability in AI-driven corrections addressing changes in physical systems. 

We believe SBAPC to be a powerful contribution in the field of model agnostic local surrogate explainability of predictive time series learning models. We also believe that this will play an important role in the ongoing competition between the different local surrogate approaches to explainable AI by overcoming LIME's instability problem \cite{visani}.

\section*{Acknowledgement}

This work has been supported by the project ‘inAIco’ (FFG-Project Nos. 862019, 3757194), the BMK, BMAW, the $\#$ upperVision2030 project SPA, and the State of Upper Austria in the frame of the SCCH competence center INTEGRATE [(FFG grant no. 892418)] as part of the FFG COMET Competence Centers for Excellent Technologies Programme.

\section*{Disclosure}
The authors have no competing interests to declare that are relevant to the content of this article.

\appendix
\section{Appendix}

\subsection{Proofs of propositions and additional technical results}\label{sec:proof}

In this section, we provide proofs of our propositions and establish some additional technical results.

\begin{proof}[Proof of proposition \ref{prop:sin2ar}]
It is clear that $y_1 = \alpha \cos(\phi)$ and $y_2 = \alpha \exp(-\beta)\cos(\omega + \phi)$. Let us take $z = -\beta + i \omega$ and $t \in \{3,\ldots,n \}$. We have
\begin{align*}
\exp(zt)
&= \exp(z(t-1))\exp(z) + \exp(z(t-1))\exp(-z-2\beta) - \exp(-2\beta)\exp(z(t-2))  \\
&= \exp(z(t-1))(\exp(z) + \exp(\bar{z})) - \exp(-2\beta)\exp(z(t-2)) \\
&= \exp(z(t-1))2\exp(-\beta)\cos(\omega) - \exp(-2\beta)\exp(z(t-2)) \\
&= \varphi_1 \exp(z(t-1)) + \varphi_2 \exp(z(t-2)).
\end{align*}
Then we obtain that
\begin{align*}
y_{t+1} 
&= \Re{\alpha \exp(i \phi)\exp(zt)} \\
&= \Re{\alpha \exp(i \phi)\qty(\varphi_1 \exp(z(t-1)) + \varphi_2 \exp(z(t-2)))} \\
&= \varphi_1\Re{\alpha \exp(i \phi)\exp(z(t-1))} + \varphi_2\Re{\alpha \exp(i \phi) \exp(z(t-2)))}\\
&= \varphi_1 y_{t-1} + \varphi_2 y_{t-2}
\end{align*}
and so the result is proved.
\end{proof}

\begin{proof}[Proof of Proposition \ref{prop:ar2sin}]
    let us denote $y(t)= \Re{\alpha \exp(i \phi)\exp(zt)}$, with $z = -\beta + i \omega$. We have $y(0)=\Re{\alpha \exp(i \phi)}=\alpha \cos \phi = y_1$ and
\begin{align*}
    y(1) &= \Re{\alpha \exp(i \phi)\exp(z)}= \alpha \exp(-\beta) \cos(\omega + \phi) \\
         &= \alpha \exp(-\beta) \cos\omega\cos\phi - \alpha \exp(-\beta) \sin\omega\sin\phi\\
         &=  y_1 \exp(-\beta) \cos\omega - y_1 \exp(-\beta) \sin\omega \tan\phi = y_2.
\end{align*}
     
    Let us show now that $y_t = y(t-1)$, $t=3,\ldots,n$, by induction on $t$. Given that $\varphi_1=2\exp(-\beta)\cos(\omega)$ and $\varphi_2=-\exp(-2\beta)$, we obtain
\begin{align*}
        \varphi_1 \exp(z) + \varphi_2
        &= 2\exp(-\beta)\cos(\omega)\exp(z)  - \exp(-2\beta)\\
        & = (\exp(z) + \exp(\bar{z}))\exp(z) - \exp(-2\beta) \\
        & = \exp(2z) + \exp(z + \bar{z}) - \exp(-2\beta)  = \exp(2 z).
\end{align*}
Therefore 
$$y_3 = \varphi_1 y_2 + \varphi_2 y_1 = \Re{\alpha \exp(i \phi)\qty(\varphi_1 \exp(z) + \varphi_2)} = \Re{\alpha \exp(i \phi)\exp(2 z)} = y(2).$$

Assuming that $y_k=y(k-1)$ for $k=1,\ldots,t$, we obtain 
\begin{align*}
 \varphi_1 \exp(z(t-1)) + \varphi_2 \exp(z(t-2)) 
 &= \exp(z(t-2))\qty(\varphi_1 \exp(z) + \varphi_2)\\
 &= \exp(z(t-2))\exp(2z) = \exp(zt),
\end{align*}
and so
\begin{align*}
y_{t+1} 
&= \varphi_1 y_t + \varphi_2 y_{t-1} = \varphi_1 y(t-1) + \varphi_2 y(t-2) \\
&= \Re{\alpha \exp(i \phi) \qty(\varphi_1 \exp(z(t-1)) + \varphi_2 \exp(z(t-2)))} \\
&= \Re{\alpha \exp(i \phi)\exp(zt)}=y(t),
\end{align*} 
which concludes the proof.
\end{proof}

\begin{lemma}\label{lem:intgrad}
    Let $f : \bbbr \times \bbbr^n \to \bbbr$ be defined as $f(x, y) = x\exp(\mu^\intercal y)$ with $\mu \in \bbbc^n$. Then for any $x,\Delta x \in\bbbr$ and $y,\Delta y \in\bbbr^n$,  it holds that
    \begin{align}
     \Delta x \int_0^1 \frac{\partial f}{\partial x} (x+h\Delta x, y+h\Delta y) \dd{h} 
     &= \Delta x \frac{\Delta g}{\mu^\intercal \Delta y} \label{alig:intgrad1}\\
     \Delta y_k \int_0^1 \frac{\partial f}{\partial y_k} (x+h\Delta x, y+h\Delta y) \dd{h} 
     &= \frac{\mu_k \Delta y_k}{\mu^\intercal \Delta y} \qty( \Delta f - \Delta x \frac{\Delta g}{\mu^\intercal \Delta y}), \label{alig:intgrad2}
    \end{align}
for every $k=1,\ldots,n$, where $\Delta g := \exp(\mu^\intercal (y+\Delta y)) - \exp(\mu^\intercal y)$ and $\Delta f := f(x+\Delta x, y+\Delta y) - f(x,y)$.
\end{lemma}
\begin{proof}
Denoting by $I$ the left-hand side of \eqref{alig:intgrad1}, we have
$$
 I  = \Delta x \int_0^1  \exp(\mu^\intercal (y+h\Delta y)) \dd{h} = \frac{\Delta x }{\mu^\intercal \Delta y} \exp(\mu^\intercal (y+h\Delta y))\Big|_0^1   = \Delta x \frac{\Delta g}{\mu^\intercal \Delta y}.
$$
Let us now denote by $J$ the left-hand side of \eqref{alig:intgrad2}. Integrating by part we obtain
\begin{align*}
 J   & = \Delta y_k \mu_k \int_0^1  (x+h\Delta x)  \exp(\mu^\intercal (y+h\Delta y)) \dd{h} \\
     & = \frac{\Delta y_k \mu_k}{\mu^\intercal \Delta y} \qty((x+h\Delta x) \exp(\mu^\intercal (y+h\Delta y)) \Big|_0^1 -  \Delta x \int_0^1  \exp(\mu^\intercal (y+h\Delta y)) \dd{h})\\
     & = \frac{\Delta y_k \mu_k}{\mu^\intercal \Delta y} \qty((x+h\Delta x) \exp(\mu^\intercal (y+h\Delta y)) \Big|_0^1 -  \frac{\Delta x}{\mu^\intercal \Delta y}  \exp(\mu^\intercal (y+h\Delta y)) \Big|_0^1)
\end{align*}
which implies \eqref{alig:intgrad2}.
\end{proof}

The following result expresses each value $y_t$ in an AR(2) process  (see \eqref{eq:ar}) as a weighted combination of the initial values $y_1$ and $y_2$, with time-varying weights $\Phi_1(t, \boldsymbol{\varphi})$ and $\Phi_2(t, \boldsymbol{\varphi})$ defined recursively in terms of a base function $\Phi(t, \boldsymbol{\varphi})$.

\begin{proposition}\label{prop:combform}
    Let $(y_t)_{t=1}^n$ be an $\mathrm{AR(2)}$ sequence with coefficients $\bvarphi=(\varphi_1, \varphi_2)$. Then
    \begin{equation*}
        y_t = \Phi_1(t, \bvarphi)y_2 +  \Phi_2(t, \bvarphi)y_1, \; t=1,\ldots,n, 
    \end{equation*}
    where $\Phi_{1}(1, \bvarphi)=0$, $\Phi_{1}(2, \bvarphi)=1$, $\Phi_{1}(t, \bvarphi)=\Phi(t-2, \bvarphi)$,  $\Phi_{2}(1, \bvarphi)=1$, $\Phi_{2}(2, \bvarphi)=0$, $\Phi_{2}(t, \bvarphi)=\varphi_2\Phi(t-3, \bvarphi)$, for $t=3,\ldots,n$, and 
    \begin{equation}\label{eq:combform}
        \Phi(t, \bvarphi) =  \sum_{k=0}^{\lfloor t/2 \rfloor } \binom{t-k}{k}\varphi_1^{t-2k}\varphi_2^k, \; t=3,\ldots,n.
    \end{equation}
\end{proposition}
\begin{proof}
    Let us denote $z_t =\bigl( \begin{smallmatrix} y_{t+1} \\ y_t\end{smallmatrix}\bigr)$,  $A = \bigl( \begin{smallmatrix} \varphi_1 & \varphi_2\\ 1 & 0\end{smallmatrix}\bigr)$ and $A^t = \bigl( \begin{smallmatrix} a^{(t)} & b^{(t)}\\ c^{(t)}  & d^{(t)} 
 \end{smallmatrix}\bigr)$. From the state space representation $z_t = A z_{t-1}$ we obtain $z_{t} = A^{t-1} z_{1}$ and so 
\begin{equation}\label{eq:arformula}  
    y_{t} = a^{(t-1)}y_2 + b^{(t-1)}y_1, \; t=3,\ldots,n.
\end{equation}
Then the result follows from \eqref{eq:arformula} and Theorem 1 in \cite{Konvalina2015}. 
\end{proof}

\begin{proposition}\label{prop:ig_formulas}
    For any $t \in \bbbr$, we have the following. 
    \begin{enumerate}
        \item If $f_{\theta}(t) = \theta^\intercal g(t)$ with $\theta \in \bbbr^q$ and $g(t)=(g_1(t),\ldots,g_q(t))$ then 
        $$
        \ig_k(f_\theta, t) = \Delta \theta_{rk} g(t), \; k=1,\ldots,q.
        $$
        
        \item If $f_\theta(t) = \alpha \exp(-\beta t) \cos(\omega t + \phi)$, with $\theta=(\alpha, \beta, \omega, \phi)$, then
        \begin{equation}\label{eq:ig}
        \ig(f_{\theta}, t) = \Re
        \begin{pmatrix} 
          \Delta \alpha_r \Delta g_r(t)\\  
        \frac{- \Delta \beta_r t}{- \Delta \beta_r t + i \Delta \omega_r t + i \Delta \phi_r} \left( \Delta f_r(t) - \Delta \alpha_r \Delta g_r(t) \right) \\
        \frac{i \Delta \omega_r t}{- \Delta \beta_r t + i \Delta \omega_r t + i \Delta \phi_r} \left( \Delta f_r(t) - \Delta \alpha_r \Delta g_r(t) \right)\\
        \frac{i \Delta \phi_r}{- \Delta \beta_r t + i \Delta \omega_r t + i \Delta \phi_r}\left( \Delta f_r(t) - \Delta \alpha_r \Delta g_r(t) \right)
        \end{pmatrix}
        \end{equation}
        where 
        $$
        \Delta g_r(t) := \frac{\exp(-\beta_0 t + i \omega_0 t + i \phi_0) -  \exp(-\beta_r t + i \omega_r t + i \phi_r)}{- \Delta \beta_r t + i \Delta \omega_r t + i \Delta \phi_r}.
        $$
        
       \item If $f_{\theta}(t) =\Phi_{1}(t, \varphi)y_2 + \Phi_{2}(t, \varphi)y_1$ with $\theta=\bvarphi=(\varphi_1,\varphi_2)$ and $\Phi_{1}, \Phi_{2}$ defined in  Proposition \ref{prop:combform}, then $\ig_1(f_\theta,0)=\ig_1(f_\theta,1) =\ig_2(f_\theta,0) =\ig_2(f_\theta,1)=0$ and, for $t=3,\ldots,n$, it hold that
       \begin{align}
        \ig_1(f_\theta,t)  &= \bar{\Phi}_{1}(t-2, 0)y_2 + \bar{\Phi}_{1}(t-3, 1)y_1 \label{aling:ig1} \\
        \ig_2(f_\theta,t)  &= \bar{\Phi}_{2}(t-2, 0)y_2 + \bar{\Phi}_{2}(t-3, 1)y_1 \label{aling:ig2},
       \end{align}
       where
        \begin{align*} 
         \bar{\Phi}_{1}(t, j)  &= \Delta \varphi_{r1}\sum_{k=0}^{\lfloor t/2 \rfloor}  \binom{t-k}{k}(t-2k) \Gamma^{t-1-2k}_{k+j},\\
         \bar{\Phi}_{2}(t, j)  &= \Delta \varphi_{r2}\sum_{k=0}^{\lfloor t/2 \rfloor}  \binom{t-k}{k}(k+j) \Gamma^{t-2k}_{k-1+j}
        \end{align*}
       and $\Gamma^m_n = \sum_{j=0}^m \sum_{k=0}^n  \binom{m}{j} \binom{n}{k} (j+k+1)^{-1} \varphi_{r1}^{m-j} \Delta \varphi_{r1}^{j} \varphi_{r2}^{n-k}  \Delta \varphi_{r2}^{k}$.
    \end{enumerate}
\end{proposition}
\begin{proof} 
\phantom{foo}
\begin{enumerate}
    \item For any $\theta' \in \bbbr^q$ we have $\frac{\partial f}{\partial \theta_k}(t, \theta') = g_k(t)$ and therefore $\ig_k(f_\theta, t) = \Delta \theta_{rk} \int_0^1 g_k(t) \dd{h} =  \Delta\theta_{rk} g_k(t)$.
    
    \item The result follows from Lemma \ref{lem:intgrad}, after projecting  \eqref{alig:intgrad1} and \eqref{alig:intgrad2} onto the real coordinate and taking $x=\alpha_r$, $\Delta x = \Delta \alpha_r$, $y=(\beta_r, \omega_r,\phi_r)$, $\Delta y=(\Delta\beta_r, \Delta\omega_r,\Delta\phi_r)$ and $\mu=(-t, it, i)$.
    
    \item For $t=0,1$ the function $\theta \mapsto f_{\theta}(t)$ is constant, therefore $\ig_1(f_\theta,0)=\ig_1(f_\theta,1) =\ig_2(f_\theta,0) =\ig_2(f_\theta,1)=0$. Let us now show the result for $t=3,\ldots,n$. Given $a,b,c,d \in \bbbr$ and $m,n \in \bbbz_+$, using the binomial formula it follows that
    \begin{equation}\label{eq:aux1}
    \int_0^1 \gamma^m(a,b,h) \gamma^n(c,d,h) \dd{h} = \sum_{j=0}^m \sum_{k=0}^n  \binom{m}{j} \binom{n}{k} \frac{a^{m-j} (b-a)^{j} c^{n-k} (d-c)^{k}}{j+k+1}. 
    \end{equation}
     Denoting $\gamma(h):=\gamma(\bvarphi_r, \bvarphi_0, h)$, we obtain from  \eqref{eq:combform} that
     \begin{equation}\label{eq:aux2} 
     \frac{\partial \Phi}{\partial \varphi_1}(t,\gamma(h))=\sum_{k=0}^{\lfloor t/2 \rfloor } \binom{t-k}{k}(t-2k)\gamma_1^{t-1-2k}(h)\gamma_2^k(h),
     \end{equation}
    and so \eqref{eq:aux1} implies, 
    \begin{equation}\label{eq:aux3}
    \int_0^1 \frac{\partial \Phi}{\partial \varphi_1}(t, \gamma( h)) \dd{h} = \bar{\Phi}_1(t, 0).
    \end{equation}
    Recall that $\Phi_{1}(t, \bvarphi)=\Phi(t-2, \bvarphi)$ by definition, so from \eqref{eq:aux3} it follows that
    \begin{equation}\label{eq:ig1}
    \int_0^1 \frac{\partial \Phi_1}{\partial \varphi_1}(t, \gamma( h)) \dd{h} = \bar{\Phi}_1(t-2, 0) 
    \end{equation}
    Similarly, given that $\Phi_{2}(t, \bvarphi)=\varphi_2\Phi(t-3, \bvarphi)$, it follows from  \eqref{eq:aux2} that
    \begin{align*}
       \frac{\partial \Phi_2}{\partial \varphi_1}(t+3, \gamma(h)) 
    &=  (\varphi_{r2} + h \Delta \varphi_{2r}) \frac{\partial  \Phi}{\partial \varphi_1}(t, \gamma(h)) \\
    &= \sum_{k=0}^{\lfloor t/2 \rfloor } \binom{t-k}{k}(t-2k)\gamma^{t-1-2k}( h)\gamma^{k+1}(h)
    \end{align*}
    and so
    \begin{equation}\label{eq:ig2}
    \int_0^1 \frac{\partial \Phi_2}{\partial \varphi_1}(t, \gamma(h)) \dd{h} = \bar{\Phi}_1(t-3, 1)
    \end{equation}
    Therefore, from \eqref{eq:ig1}  and \eqref{eq:ig2}, we obtain \eqref{aling:ig1}. The proof of \eqref{aling:ig2} follows from similar arguments. 

    \end{enumerate}
\end{proof}

\subsection{ODE governed time-series}\label{sec:ode}

The time series displayed at the first row of Figure \ref{fig:synthetic1} and at Figure \ref{fig:fm1} correspond to discrete observations of the form
$$y_t = u(t-1), \;t = 1, \ldots, n$$
where  $u : \bbbr \to \bbbr$ is the solution of a non-homogeneous initial value problem (IVP), or one with time-dependent coefficients; see Table \ref{tab:ivp} and Proposition \ref{prop:ivp}.

\begin{table}[ht]
    \centering
    \caption{Synthetic data description.}
    \resizebox{\textwidth}{!}{
    \begin{tabular}{|l|l|c|c|c|c|}
        \hline
        \textbf{Data} & $\boldsymbol{u(t)}$ & $\boldsymbol{u_0}$ & $\boldsymbol{v_0}$ & \textbf{Param.} & $\boldsymbol{t^*}$ \\ \hline
        Step   & $u_0  + FH(t-t^*)$     & $-1$  &   - &   $F=2$  & 48.5  \\ 
        Ramp   & $u_0 + v_0 t + F H(t-t^*)$ &  $23.5$ & -1 &   $F=2$ & 48.5\\ 
        SinACP   & $u_0 \cos(\omega t) + H(t-t^*)\qty(\flatfrac{F}{w})\sin(\omega(t-t^*))$  & 1 & 0 & \makecell{$\omega= \flatfrac{2\pi}{24}$ \\ $F=-\omega$} & 55\\ 
        SinFCP   & $\begin{cases} u_0\cos(\omega t) \quad \text{if } t <  t^* \\ u_0\cos(\nu t + (\omega - \nu) t^*) \quad \text{if } t \geq t^* \end{cases}$  & 1 & 0 & \makecell{$\omega=\flatfrac{2\pi}{40}$ \\ $\nu = 2 \omega$} & 81\\ \hline
    \end{tabular}
    }
    \label{tab:ivp}
\end{table}

These data represent discrete observations of the one-dimensional position of an object (or another one-dimensional dynamical system) governed by an ordinary differential equation (ODE). In the cases of Step, Ramp, and SinACP, these observations correspond to scenarios where an external force $F \in \mathbb{R}$ acts at a specific time $t^* \in \mathbb{R}$ for a very short duration, causing a significant change in the object's position or velocity \cite{Brooks17}. Mathematically, such external forces are modeled as $F \delta(\cdot - t^*)$, where $\delta$ denotes the Dirac delta distribution. 

Examples of such scenarios include hitting a nail with a hammer (Step), kicking an oncoming football in the opposite direction (Ramp), and a mass attached to a string subjected to an impulsive external force (SinACP). In the case of SinFCP, the observed data correspond to a situation where a mass attached to a string undergoes a sudden loss of mass and/or an increase in string stiffness.

In what follows, we show that the functions $u$ in Table \ref{tab:ivp} are solutions to specific IVP formulated within the framework of distribution theory. For a detailed theoretical background on the distribution theory applied in this paper, we refer the reader to \cite{duistermaat2010distributions}.

\begin{lemma}\label{lem:fourier1}
    Let $\ft$ be the Fourier transform over distributions, $H$ the Heaviside function on the real line and $\omega>0$, then we have the formulas:
    \begin{enumerate}
        \item $\ft \qty {\sin (\omega \cdot)}(\xi) = -i \pi (\delta(\xi-\omega) - \delta(\xi+\omega))$ \label{lem:1}
        \item $\ft \qty {\cos (\omega \cdot)}(\xi) =  \pi (\delta(\xi-\omega) +  \pi \delta(\xi+\omega))$ \label{lem:2}
        \item $\ft \qty {H\sin (\omega \cdot)}(\xi) = \frac{1}{2}\ft \qty {\sin (\omega \cdot)}(\xi) + \frac{\omega}{\omega^2 - \xi^2}$ \label{lem:3}
        \item $\ft \qty {H\cos (\omega \cdot)}(\xi) = \frac{1}{2}\ft \qty {\cos (\omega \cdot)}(\xi) + \frac{i\xi}{\omega^2-\xi^2}$ \label{lem:4}
    \end{enumerate}
\end{lemma}
\begin{proof}
    These formulas are well known, see for instance \cite{strichartz1994}.
\end{proof}

\begin{lemma}\label{lem:fourier2}
    Given $\omega>0$ we have
    \begin{enumerate}
        \item $\abs{t}''=2 \delta(t)$ \label{lem:5}
        \item $(H(t)\sin(\omega t))'' = H(t)\omega^2\sin(\omega t) - \omega \delta(t)$ \label{lem:6}
        \item $(H(t)\cos(\omega t))'' = H(t)\omega^2\cos(\omega t) - \delta'(t)$ \label{lem:7}
    \end{enumerate}
\end{lemma}
\begin{proof}
\phantom{foo}
    \begin{enumerate}
        \item See Problem 4.1 in \cite{duistermaat2010distributions}.
        \item This result is proved in Problem 9.14 in  \cite{duistermaat2010distributions}, here we provide an alternative proof. Recall that for a given distribution $v$ it holds that
        \begin{equation}\label{eq:fouriera}
            v''(t) = \ft^{-1} \qty{ \xi^2 \ft \qty{v}(\xi) }.
        \end{equation}
        Now, from Lemma \ref{lem:fourier1} part \ref{lem:3}, we have
        \begin{equation}\label{eq:fourierb}
        \xi^2 \ft \qty {H\sin (\omega \cdot)}(\xi) = \xi^2\frac{1}{2}\ft \qty {\sin (\omega \cdot)}(\xi) + \frac{\xi^2\omega}{\omega^2 - \xi^2}. 
        \end{equation}
        Therefore, from \eqref{eq:fouriera} and \eqref{eq:fourierb} we obtain
        \begin{equation}\label{eq:fourierc}
            (H(t)\sin(\omega t))'' = \frac{1}{2}\omega^2 \sin(wt) + \ft^{-1}\qty{\frac{\xi^2\omega}{\omega^2 - \xi^2}}(t).
        \end{equation}
        Also, it can be shown that 
        \begin{equation}\label{eq:fourierd}
        \ft^{-1}\qty{\frac{\xi^2\omega}{\omega^2 - \xi^2}}(t) = \frac{1}{2} \sgn(t) \omega^2  \sin(wt) - \omega \delta(t),
        \end{equation}
        and so the result follows by combining \eqref{eq:fourierc} and \eqref{eq:fourierd}. 
        \item The result is obtained by following the steps of the part 2 and proceeding analogously. Instead of \eqref{eq:fourierd} we now use the formula
        \begin{equation*}
        \ft^{-1}\qty{\frac{i\xi^3}{\omega^2 - \xi^2}}(t) = \frac{1}{2} \sgn(t) \omega^2  \cos(wt) -  \delta'(t).
        \end{equation*}
    \end{enumerate}
\end{proof}

\begin{proposition}\label{prop:ivp}
For $u_0,v_0,F \in \bbbr$ and $t^*,\omega,\nu > 0$, we have the following initial value problems (IVP) solutions:
\begin{enumerate}
    \item The distribution 
    \begin{equation}\label{eq:solstep}
    u(t) = u_0 + F H(t - t^*),
    \end{equation}
    is solution of the IVP $u'(t) = F\delta(t - t^*)$,  $u(0)=u_0$, where $H$ denotes the Heaviside function.
    \item The distribution 
    \begin{equation}\label{eq:solramp}
    u(t) = u_0 + v_0 t + FH(t-t^*), 
    \end{equation}
    is solution of the IVP $u''(t) = F\delta(t - t^*)$, $u(0)=u_0 $, $u'(0)=v_0$. 
    \item The distribution 
    \begin{equation}\label{eq:solsin}
    u(t) = u_0 \cos(\omega t) + H(t-t^*)\qty(\flatfrac{F}{w})\sin(\omega(t-t^*)) , 
    \end{equation}
    is solution of the IVP $u''(t) + \omega^2 u(t) = F\delta(t - t^*)$, $u(0)=u_0$, $u'(0)=0$.
    \item The distribution
    \begin{equation}\label{eq:solfreq}
    u(t) = 
    \begin{cases}
        u_0\cos(\omega t) \quad \text{if } t \leq  t^* \\
        u_0\cos(\nu t + (\omega - \nu) t^*) \quad \text{if } t > t^*
    \end{cases}
    \end{equation}
    is solution of the IVP $u''(t) + \qty(\omega^2 + H(t-t^*)(\nu^2 - \omega^2)) u(t) = 0$, $u(0)=u_0$, $u'(0)=0$.
\end{enumerate}
\end{proposition}
% u(t) = (1-H(t-t^*)) u_0\cos(\omega t) + u_0 H(t-t^*)\cos(\nu t + (\omega - \nu) t^*)
\begin{proof} 
\phantom{foo}
    \begin{enumerate}
        \item It is a direct consequence from the fact that $u_h:=u_0$ satisfies the homogeneous equation and that $H' = \delta$ (see for instance Example 4.2 in \cite{duistermaat2010distributions}).
        \item We have that the distribution $u_0 + v_0 t + \flatfrac{F}{2} (t-t^*)$ satisfies the homogeneous equation and that $\abs{\cdot}''= 2 \delta$ (see Lemma \ref{lem:fourier2} part \ref{lem:1}), which implies the result.
        \item It is easy to check that  $u_h = u_0 \cos(\omega t)$ is a solution of the homogeneous equation. Also, from Lemma \ref{lem:fourier2} part \ref{lem:6}, it follows that the distribution $u_p:=H(t) (\flatfrac{F}{\omega}) \sin(\omega(t))$ satisfies $u_p'' + \omega^2 u_p = F \delta$. Therefore $u=u_h + u_p(t-t^*)$ satisfies the IVP.
        \item For simplicity, let us prove the result for $t^*=0$ and $u_0=1$, the general case follows from analogous argument. In this case,  $u(t) = \cos(\omega t) - H(t) \cos(\omega t) + H(t)\cos(\nu t)$ and so the result follows from Lema \ref{lem:fourier2} part \ref{lem:7}. 
    \end{enumerate}
\end{proof}

\printbibliography

\end{document}